%% file: main.tex
\documentclass[10pt,twocolumn,letterpaper]{article}

\usepackage[pagenumbers]{wacv} 

\usepackage{graphicx}
\usepackage{amsmath}
\usepackage{amssymb}
\usepackage{booktabs}

\usepackage[dvipsnames]{xcolor}
\usepackage{multirow}
\usepackage{colortbl}
\usepackage{comment}
\usepackage{times}
\usepackage{epsfig}
\usepackage{xcolor}
\usepackage{boldline}
\usepackage{algpseudocode}
\usepackage{algorithm}

\definecolor{wacvblue}{rgb}{0.21,0.49,0.74}
\usepackage[pagebackref,breaklinks,colorlinks,allcolors=wacvblue]{hyperref}

\def\wacvPaperID{406} 
\def\confName{WACV}
\def\confYear{2026}

\title{AdaptViG: Adaptive Vision GNN with Exponential Decay Gating}

\author{Mustafa Munir\\
The University of Texas at Austin\\
{\tt\small mmunir@utexas.edu}
\and
Md Mostafijur Rahman\\
The University of Texas at Austin\\
{\tt\small mostafijur.rahman@utexas.edu }
\and
Radu Marculescu\\
The University of Texas at Austin\\
{\tt\small radum@utexas.edu} \\
}
\begin{document}
\maketitle

\begin{abstract}
Vision Graph Neural Networks (ViGs) offer a new direction for advancements in vision architectures. While powerful, ViGs often face substantial computational challenges stemming from their graph construction phase, which can hinder their efficiency. To address this issue we propose AdaptViG, an efficient and powerful hybrid Vision GNN that introduces a novel graph construction mechanism called Adaptive Graph Convolution. This mechanism builds upon a highly efficient static axial scaffold and a dynamic, content-aware gating strategy called Exponential Decay Gating. This gating mechanism selectively weighs long-range connections based on feature similarity. Furthermore, AdaptViG employs a hybrid strategy, utilizing our efficient gating mechanism in the early stages and a full Global Attention block in the final stage for maximum feature aggregation. Our method achieves a new state-of-the-art trade-off between accuracy and efficiency among Vision GNNs. For instance, our AdaptViG-M achieves 82.6\% top-1 accuracy, outperforming ViG-B by 0.3\% while using 80\% fewer parameters and 84\% fewer GMACs. On downstream tasks, AdaptViG-M obtains 45.8 mIoU, 44.8 APbox, and 41.1 APmask, surpassing the much larger EfficientFormer-L7 by 0.7 mIoU, 2.2 APbox, and 2.1 APmask, respectively, with 78\% fewer parameters.
\end{abstract}

\vspace{-3mm}

\section{Introduction}
\label{sec:intro}

\begin{figure*}[t]
\centering
\begin{subfigure}[b]{0.43\textwidth}
   \includegraphics[width=\textwidth]{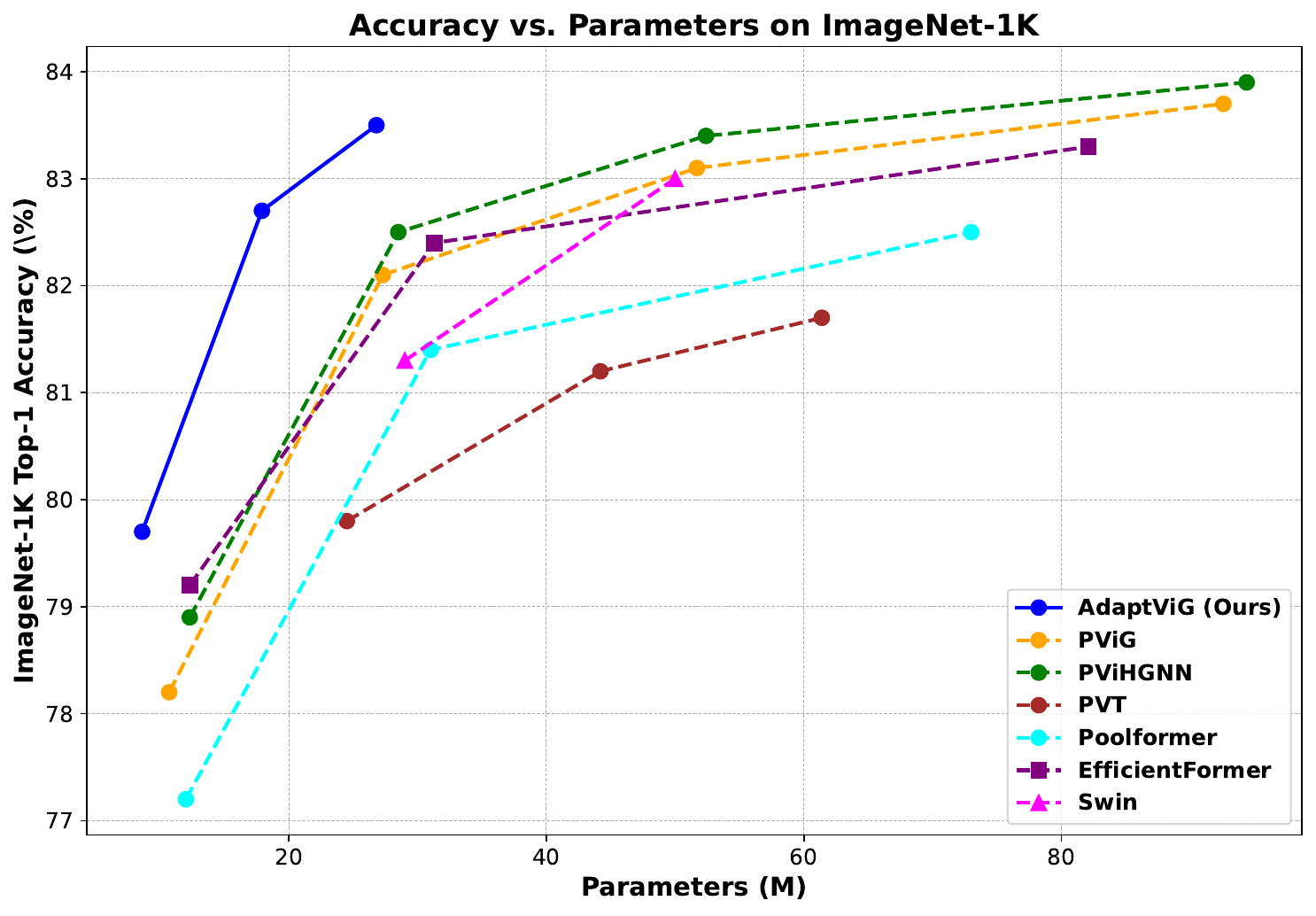}
   \caption{Accuracy vs. Parameters (M)}
   \label{fig:pareto_params}
\end{subfigure}
\hfill 
\begin{subfigure}[b]{0.43\textwidth}
   \includegraphics[width=\textwidth]{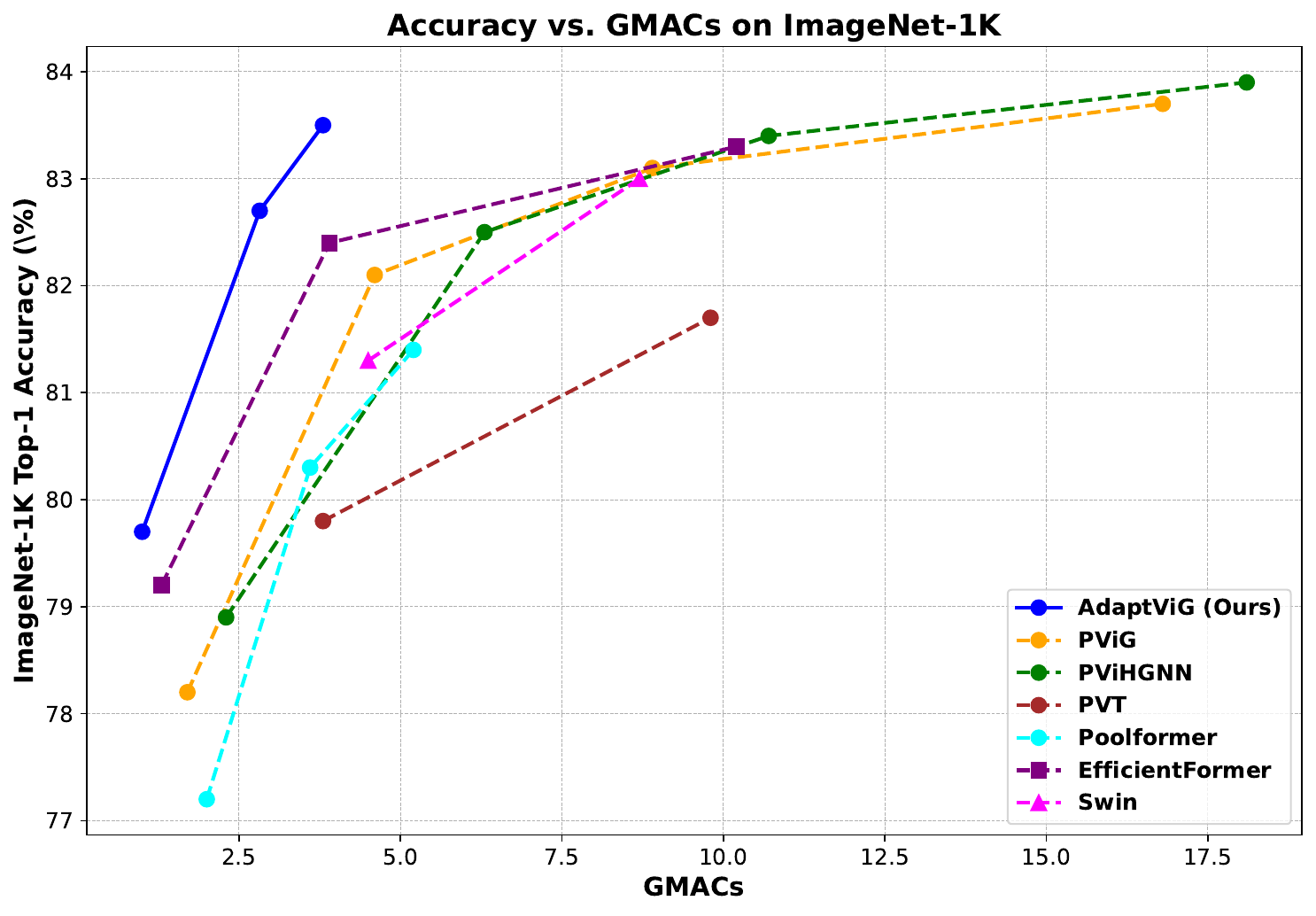}
   \caption{Accuracy vs. GMACs}
   \label{fig:pareto_gmacs}
\end{subfigure}
\caption{\textbf{Comparison of model efficiency and performance on ImageNet-1K.} AdaptViG establishes a new state-of-the-art Pareto frontier, achieving higher top-1 accuracy than competing models with (a) a lower number of parameters and (b) a lower number of GMACs.}
\label{fig:pareto_curves}
\vspace{-2mm}
\end{figure*}

The success of deep learning in computer vision, driven by Convolutional Neural Networks (CNNs)~\cite{Resnet, mobileone2022, MobileNet, MobileNetv2, mehta2018espnet} and Vision Transformers (ViTs)~\cite{ViT, liu2021swin, vaswani2017attention, MobileFormer}, has recently been challenged by a new paradigm: Vision Graph Neural Networks (ViGs)~\cite{Vision_GNN, huang2023vigu, GreedyViG}. In contrast to the rigid grids of CNNs or sequential processing of ViTs, ViGs offer a more flexible image representation by modeling patches as \textit{nodes} connected by \textit{edges} in a graph, making them a natural fit for capturing complex spatial relationships and the irregular shapes of objects.

Despite their promise, the practical application of ViGs is hindered by the graph construction stage. The pioneering ViG model relies on a k-Nearest Neighbor (KNN) search~\cite{Vision_GNN, KNN}, an approach that scales quadratically with the number of patches and is thus computationally prohibitive for many use cases. This has led to an increase in research focused on creating more efficient graph construction policies, but existing methods face a number of key limitations:

\begin{enumerate}
    \item \textbf{The Efficiency vs. Adaptability Trade-off:} To solve the KNN bottleneck, static methods like MobileViG \cite{MobileViG, MobileViGv2, LogViG} use fixed axial connection patterns that are highly efficient, but they sacrifice the crucial ability to adapt the graph structure to the image content.
    \item \textbf{The Locality vs. Globality Trade-off:} Other approaches like WiGNet~\cite{spadaro2025wignet} partitions the image into rigid windows and builds graphs locally, while ClusterViG~\cite{parikh2025clustervig} uses content-based clustering to form partitions. While efficient, these methods limit the receptive field by not forming direct, long-range connections across the entire image.
\end{enumerate}

In this work, we introduce \textbf{AdaptViG}, a novel hybrid Vision GNN architecture designed to resolve this efficiency versus adaptability dilemma. The core of our model is the proposed \textbf{Adaptive Graph Convolution (AGC)}, which features a numerically stable and content-aware \textbf{Exponential Decay Gating} mechanism. Our method builds upon an efficient static scaffold of local and logarithmic connections and then dynamically modulates the strength of long-range links. This decision is governed by a learned, per-edge gating function derived from the feature similarity between nodes, calculated using an exponential decay function. This promotes a sparse, meaningful graph structure tailored to each image's contents while ensuring robust training. The effectiveness of our approach is demonstrated in \Cref{fig:pareto_curves}, where \textbf{AdaptViG} establishes a new state-of-the-art Pareto frontier. For instance, our \textbf{AdaptViG-M} model achieves 82.6\% top-1 accuracy, outperforming strong baselines like ViG-B by 0.3\% while using 80\% fewer parameters.

Our contributions are summarized as follows:
\begin{itemize}
    \item We propose \textbf{Adaptive Graph Convolution (AGC)}, a novel graph construction algorithm that builds on an efficient static scaffold of local and logarithmic connections and then dynamically modulates the strength of long-range links based on learned, per-edge feature similarity.
    \item We propose \textbf{AdaptViG}, a hierarchical hybrid Vision GNN that leverages our efficient gating mechanism in early, high-resolution stages and a powerful Global Attention mixer in the final stage for a superior accuracy-efficiency trade-off.
    \item We demonstrate through extensive experiments that \textbf{AdaptViG} achieves a state-of-the-art balance of accuracy, parameters, and GMACs on ImageNet-1K \cite{imagenet1k} classification and key downstream tasks.
\end{itemize}

The rest of this paper is organized as follows. Section~\ref{sec:related_work} covers related work in the CNN, ViT, and ViG space. Section~\ref{sec:methodology} describes the new design methodology behind our AGC and the \textbf{AdaptViG} architecture. Section~\ref{sec:experiments} details the experimental setup and results, and Section~\ref{sec:conclusion} summarizes our main contributions.


\vspace{-2mm}

\section{Related Work}
\label{sec:related_work}

\vspace{-2mm}

Vision Graph Neural Networks (ViGs) have recently emerged as a compelling alternative to convolutional neural networks (CNNs), vision transformers (ViTs), and vision state space models (SSMs). This section reviews these architectural families, focusing on the evolution within ViGs that motivates our work.

\vspace{-1mm}

\subsection{Vision Architectures}

\vspace{-1mm}

\noindent
\textbf{Convolutional Neural Networks (CNNs)} have been the foundation of modern computer vision, with seminal works like AlexNet \cite{Alexnet2012}, ResNet~\cite{Resnet}, and VGGNet~\cite{VGGNet} establishing deep learning as the state-of-the-art. Their success is rooted in strong inductive biases like locality and weight sharing, which are highly effective for learning hierarchical visual features. Significant research has also been dedicated to efficient CNN Architectures like MobileNet~\cite{MobileNet, MobileNetv2} and EfficientNet~\cite{tan2019efficientnet, tan2021efficientnetv2}, which introduced innovations such as depthwise separable convolutions and optimized scaling strategies. While powerful for local feature extraction, the inherently local receptive field of CNNs can be a limitation for capturing the global context. More recently, this includes gated CNNs like HorNet~\cite{HorNet} and MogaNet~\cite{MogaNet}, which introduce dynamic mechanisms to control information flow and model high-order spatial interactions.

\vspace{1mm}
\noindent
\textbf{Vision Transformers (ViTs)} initiated a paradigm shift by adapting the Transformer architecture~\cite{vaswani2017attention}, originally from natural language processing, for vision tasks. The foundational ViT~\cite{ViT} model treats an image as a sequence of patches and leverages self-attention to model global relationships between them, achieving excellent performance. However, the quadratic complexity of self-attention with respect to the number of patches presents a major computational hurdle, especially for high-resolution images. This has led to the development of more efficient variants. For instance, Swin Transformer~\cite{liu2021swin} computes attention within shifted local windows, while hybrid models like MobileViT~\cite{MobileViT, MobileViTv2} and EfficientFormer~\cite{EfficientFormer, EfficientFormerv2} merge convolutional and attention mechanisms to create more efficient backbones. MaxViT~\cite{MaxViT} and CrossViT \cite{CrossViT} also enhance efficiency through mixing local and global attention \cite{MaxViT} and processing multi-scale patches \cite{CrossViT} respectively.

\vspace{1mm}
\noindent
\textbf{State Space Models (SSMs)} have also recently emerged as a competing paradigm. Inspired by Mamba~\cite{Mamba}, models like ViM~\cite{Vim} and VMamba~\cite{VMamba} adapt SSMs for vision, offering linear complexity for modeling long-range dependencies, presenting a compelling alternative to both Transformers and GNNs.

\vspace{1mm}
\noindent
\textbf{Vision Graph Neural Networks (ViGs)} represent images as graphs of interconnected nodes (patches), offering a more flexible structure than the rigid grids of CNNs \cite{chen2024survey, jiao2022graph}. The pioneering Vision GNN (ViG)~\cite{Vision_GNN} demonstrated the viability of this approach for general vision tasks, using the k-Nearest Neighbors (KNN) \cite{KNN} algorithm to connect semantically similar nodes. This KNN algorithm, however, introduces a significant computational bottleneck due to its quadratic scaling with the number of nodes. Addressing this challenge has become a central theme in subsequent ViG research, leading to several distinct approaches:

\begin{itemize}
    \item \textbf{Static Graphs:} To eliminate the KNN bottleneck, new methods \cite{MobileViG, MobileViGv2, LogViG} employ fixed, efficient connection patterns that are content-agnostic. While highly efficient, these static graph topologies sacrifice the key advantage of graphs: the ability to adapt their structure to the input image contents.

    \item \textbf{Localized Graphs:} To improve scalability while retaining some dynamism, other methods localize graph construction. WiGNet~\cite{spadaro2025wignet} partitions the image into windows and builds graphs locally within them, while ClusterViG~\cite{parikh2025clustervig} forms partitions via content-based clustering. These approaches reduce complexity, but restrict the formation of direct, long-range connections across the entire image in the initial stages.
\end{itemize}

AdaptViG is designed to address these limitations of prior art. More precisely, our approach "marries" the efficiency of a static graph scaffold with a robust and dynamic soft masking strategy. Our proposed \textbf{Exponential Decay Gating} operates on a per-edge, per-image basis, offering a stable and highly adaptive method to construct a sparse, meaningful graph, thereby resolving the critical trade-off between adaptability and efficiency in Vision GNNs.


\vspace{-2mm}

\section{Methodology*}
\label{sec:methodology}
\def\thefootnote{*}\footnotetext{For the detailed theoretical basis of our AGC, including the gradient flow analysis, please see the Supplementary Materials Section \ref{app:theory}.}

\vspace{-2mm}

In this section, we present the design of our proposed AdaptViG model. Section~\ref{ssec:gating_mechanism} first introduces Adaptive Graph Convolution (AGC) and the Exponential Decay Gating mechanism, which forms the basis of our graph construction. Section~\ref{ssec:graph_structure} presents a structural analysis of the graphs generated by our method to justify our design. Section~\ref{ssec:global_attention} then describes the global attention mechanism used as a powerful feature aggregator in the final stage of our model. Section~\ref{ssec:graph_block} explains how these mechanisms are integrated into the AGC and Attention Blocks. Finally, Section~\ref{ssec:architecture} details the overall architecture of AdaptViG to achieve a new state-of-the-art efficiency-accuracy trade-off.

\begin{figure}[t]
\centering
\includegraphics[width=0.9\columnwidth]{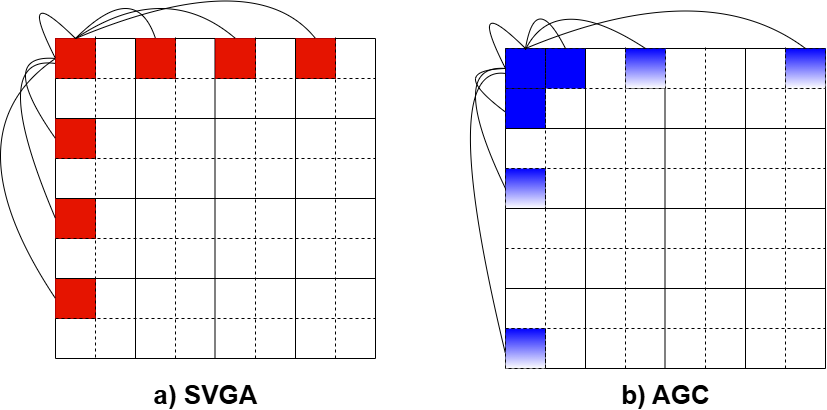}
\caption{\textbf{Comparison of SVGA vs. our AGC.} \textbf{a)} SVGA establishes fixed, unweighted connections to nodes based on a fixed number of hops. \textbf{b)} Our proposed AGC creates immediate local connections along with content-aware distant connections where the strength is weighted by feature similarity using Exponential Decay Gating, visualized by the intensity of the blue color.}
\label{fig:adapt_conv_block}
\end{figure}

\vspace{-1mm}

\subsection{Adaptive Graph Convolution (AGC)}
\label{ssec:gating_mechanism}
We propose \textbf{Adaptive Graph Convolution}, an efficient dynamic graph construction algorithm that avoids the computational expense of KNN~\cite{Vision_GNN} and the information loss caused by the hard pruning of previous dynamic methods~\cite{GreedyViG}. Our approach builds upon an efficient static scaffold of axial and logarithmic connections ~\cite{MobileViG, LogViG} and introduces a dynamic, per-edge gating function to weigh connections based on feature similarity. This creates a sparse, content-aware graph that adapts to each input image.

Our gating mechanism called Exponential Decay Gating is designed to be intrinsically stable and granular. For a given node $x_i$ and a potential long-range neighbor $x_n$, we first calculate the L1 distance, $d_{in} = ||x_i - x_n||_1$, to measure their feature dissimilarity. We then compute a "soft" gating value using an exponential decay function:

\vspace{-2mm}

\begin{equation}
\label{eq:gating}
g_{in} = \exp\left(-\frac{d_{in}}{T}\right)
\end{equation}

\vspace{-2mm}

Here, $T$ is a learnable (positive) scalar temperature parameter that controls the sensitivity of the gate. A smaller $T$ results in a sharper decay, making the gate more selective. This function smoothly maps the distance to a gate value in the range $(0, 1]$, where a value of 1 signifies identical patches (zero distance) and values approaching 0 signify highly dissimilar patches. This gating value is then used to modulate the information flow from the neighbor, as detailed in Algorithm~\ref{alg:adaptconv}.

Operationally, the AGC module first establishes a static scaffold of connections for an input feature map $X \in \mathbb{R}^{B \times C \times H \times W}$. This scaffold includes both guaranteed local connections at a fixed distance $K$ and potential long-range connections at exponentially increasing distances ($2^1, 2^2, \dots$) along each axis. For each potential connection between a node $x_i$ and a neighbor $x_n$, the feature relationship is captured as a max-relative difference \cite{maxrel}, i.e.,  $max(x_n - x_i)$. Our Exponential Decay Gate, $g_{in}$, is applied as a soft multiplier to this difference for all long-range connections, effectively weighing their importance based on contents similarity. The final aggregated feature for each node, $X'_j$, is the result of taking the element-wise maximum over all local and gated long-range neighbor features. This entire process is differentiable, numerically stable, and efficient as it relies on simple tensor-shifting operations.

More precisely the key advantages of this approach are:
\begin{itemize}
    \item \textbf{Numerical Stability:} By design the gating function is division-free with respect to any feature-dependent values, avoiding division-by-zero errors that cause training failures in other methods.
    \item \textbf{Content-Aware Sparsity:} The mechanism naturally weighs down connections between semantically dissimilar patches on a per-image basis, creating a sparse and meaningful graph without using a hard threshold.
    \item \textbf{Efficiency:} The process builds on a fast axial scaffold using efficient `roll` operations and simple, parallelizable per-edge calculations, adding minimal computational overhead.
\end{itemize}

\begin{algorithm}
\caption{Adaptive Graph Convolution (AGC)}
\footnotesize
\label{alg:adaptconv}
\begin{algorithmic}[1]
\State \textbf{Input:} Feature map $X \in \mathbb{R}^{B \times C \times H \times W}$, Local distance $K$, Temperature $T$
\State \textbf{Output:} Aggregated feature map $X_{out} \in \mathbb{R}^{B \times C \times H \times W}$

\State Initialize max-relative feature $X_j \leftarrow \mathbf{0} \in \mathbb{R}^{B \times C \times H \times W}$

\State $X_{j} \leftarrow \max(X_{j}, \text{roll}(X, -K, 2) - X)$
\State $X_{j} \leftarrow \max(X_{j}, \text{roll}(X, -K, 3) - X)$

\For{$i$ in $1 \dots \lfloor\log_2 H\rfloor$}
    \State $X_N \leftarrow \text{roll}(X, -2^i, 2)$
    \State $D \leftarrow ||X - X_N||_1$ \Comment{Compute L1 distance map}
    \State $G \leftarrow \exp(-D / (|T| + \epsilon))$ \Comment{Compute gate map}
    \State $X_j \leftarrow \max(X_j, G \odot (X_N - X))$\Comment{$\odot$=Element-wise prod.}
\EndFor
\For{$i$ in $1 \dots \lfloor\log_2 W\rfloor$}
    \State $X_N \leftarrow \text{roll}(X, -2^i, 3)$
    \State $D \leftarrow ||X - X_N||_1$
    \State $G \leftarrow \exp(-D / (|T| + \epsilon))$
    \State $X_j \leftarrow \max(X_j, G \odot (X_N - X))$
\EndFor

\State $X_{cat} \leftarrow \text{concat}([X, X_j], \text{dim}=1)$
\State $X_{out} \leftarrow \text{ConvBlock}(X_{cat})$
\State \textbf{return} $X_{out}$
\end{algorithmic}
\end{algorithm}

\vspace{-4mm}

\subsection{Graph Structure Analysis}
\label{ssec:graph_structure}

\vspace{-2mm}

To provide theoretical support for our design, we analyze the topological properties of the graphs generated by AGC. An ideal graph for vision tasks should facilitate both robust local feature aggregation and efficient global information propagation. To determine the efficacy of the graph construction we evaluate two key metrics: the Average Clustering Coefficient (C) \cite{watts1998collective} and the Spectral Gap (S) \cite{spectral_gap_eigenvalues}.

The \textbf{Average Clustering Coefficient (C)} reflects local connectivity and is the average of the local clustering coefficient $C_i$ over all $N$ nodes in the graph:

\vspace{-2mm}

\begin{equation}
    C = \frac{1}{N} \sum_{i=1}^{N} C_i, \quad \text{where} \quad C_i = \frac{2 E_i}{k_i(k_i - 1)}
\end{equation}

\vspace{-2mm}

Here, $k_i$ is the degree (number of neighbors) of node $i$, and $E_i$ is the number of edges between those $k_i$ neighbors. A higher value indicates a more connected local structure.

The \textbf{Spectral Gap (S)} indicates a graph's expansion properties and is the difference between the two largest eigenvalues ($\lambda_1, \lambda_2$) of the graph's adjacency matrix $A$:

\vspace{-2mm}

\begin{equation}
    S = \lambda_1 - \lambda_2
\end{equation}

\vspace{-2mm}

A larger spectral gap signifies that information can mix more rapidly across the graph. We compare our AGC against static (LSGC) and dynamic (KNN) methods, with results presented in Table~\ref{tab:graph_analysis}.

\begin{table*}[t!]
	\centering
	\caption{\textbf{Graph property analysis across different methods and feature map resolutions.} We report Avg. Clustering Coefficient (C) and Spectral Gap (S). Arrows indicate the desired direction for each metric. Best result for each metric is in \textbf{bold}.}
	\label{tab:graph_analysis}
	\resizebox{0.8\textwidth}{!}{%
	\begin{tabular}{l|cc|cc|cc|cc}
		\toprule
		& \multicolumn{2}{c|}{\textbf{$56 \times 56$}} & \multicolumn{2}{c|}{\textbf{$28 \times 28$}} & \multicolumn{2}{c|}{\textbf{$14 \times 14$}} & \multicolumn{2}{c}{\textbf{$7 \times 7$}} \\
		\cmidrule(lr){2-3} \cmidrule(lr){4-5} \cmidrule(lr){6-7} \cmidrule(lr){8-9}
		\textbf{Method} & C $\uparrow$ & S $\uparrow$ & C $\uparrow$ & S $\uparrow$ & C $\uparrow$ & S $\uparrow$ & C $\uparrow$ & S $\uparrow$ \\
		\midrule
		LSGC \cite{LogViG} & 0.213 & 0.229 & 0.224 & 0.310 & 0.248 & 0.589 & 0.293 & 0.954 \\
		k-Nearest Neighbors \cite{Vision_GNN} & 0.157 & 0.854 & 0.201 & 0.921 & 0.288 & 1.153 & 0.412 & 1.392 \\
		\midrule
		\textbf{Adaptive Graph Convolution} & \textbf{0.523} & \textbf{1.024} & \textbf{0.519} & \textbf{1.215} & \textbf{0.507} & \textbf{1.560} & \textbf{0.492} & \textbf{1.887} \\
		\bottomrule
	\end{tabular}
	}
    \vspace{-3mm}
\end{table*}

The analysis in Table~\ref{tab:graph_analysis} reveals that our Adaptive Graph Convolution (AGC) constructs a topology that is structurally superior to baseline static and dynamic methods across all key metrics. For capturing the local structure, a high clustering coefficient is a key component of \textbf{small-world networks}~\cite{watts1998collective}, enabling robust local feature aggregation. Our AGC consistently achieves the highest clustering coefficient across all resolutions, surpassing Logarithmic Scalable Graph Construction \cite{LogViG} (e.g., 0.523 vs. 0.213 at $56 \times 56$ resolution). This demonstrates its ability to form strong local connections.

For capturing the global structure, the spectral gap indicates a graph's efficiency at information mixing \cite{spectral_gap_eigenvalues}, a hallmark of \textbf{Ramanujan-like expander graphs}~\cite{lubotzky1988ramanujan}. Here again, our AGC obtains the largest spectral gap, significantly outperforming other methods. For instance, at a $14 \times 14$ resolution, AGC achieves a spectral gap of 1.560, a 35\% improvement over KNN (1.153). By simultaneously maximizing both local clustering and the spectral gap for global expansion, our method creates a graph that is theoretically better suited for the complex demands of visual recognition tasks.

\vspace{-1mm}

\subsection{Global Attention as a Final Stage Mixer}
\label{ssec:global_attention}

\vspace{-1mm}

In the final stage of AdaptViG, where the spatial resolution of the feature map is at its smallest (e.g., 7$\times$7), the computational cost of dense attention becomes manageable. At this point, achieving maximum feature aggregation is more critical for classification than the efficiency of sparse connections. Therefore, we replace our efficient AGC with an Attention block which implements scaled dot-product attention. This hybrid strategy ensures that our model is both efficient in high-resolution early stages and maximally expressive in the final, low-resolution stage.

The process for the new Attention block is as follows. Given an input feature map $X \in \mathbb{R}^{B \times C \times H \times W}$, we first project it into Query (Q), Key (K), and Value (V) tensors using 1$\times$1 convolutions:
\begin{equation}
Q = \phi_q(X), \quad K = \phi_k(X), \quad V = \phi_v(X)
\end{equation}
where $\phi$ represents a 1$\times$1 convolution. The tensors are then flattened from shape $(B, C', H, W)$ to $(B, N, C')$, where $N = H \times W$ is the number of tokens (patches) and $C'$ is the channel dimension for Q, K, and V. To enhance training stability, we apply Layer Normalization \cite{LayerNorm} to the Query and Key tensors before the attention operation:

\vspace{-4mm}
\begin{equation}
Q_{\text{norm}} = \text{LayerNorm}(Q), \quad K_{\text{norm}} = \text{LayerNorm}(K)
\end{equation}
The attention map is then computed using the scaled dot-product attention formula:
\vspace{-2mm}
\begin{equation}
\text{Attention}(Q, K, V) = \text{softmax}\left(\frac{Q_{\text{norm}} K_{\text{norm}}^T}{\sqrt{d_k}}\right)V
\end{equation}
where $d_k$ is the dimension of the key vectors. The output of the attention mechanism, which represents the aggregated features, is then reshaped back to the original 2D feature map dimensions, $X_{\text{attn}} \in \mathbb{R}^{B \times C \times H \times W}$. This feature map is then fused with the original input $X$ using a max-relative difference \cite{maxrel} and concatenated before being passed through a final projection layer.

\begin{figure}[h]
    \centering
    \includegraphics[width=1.0\columnwidth]{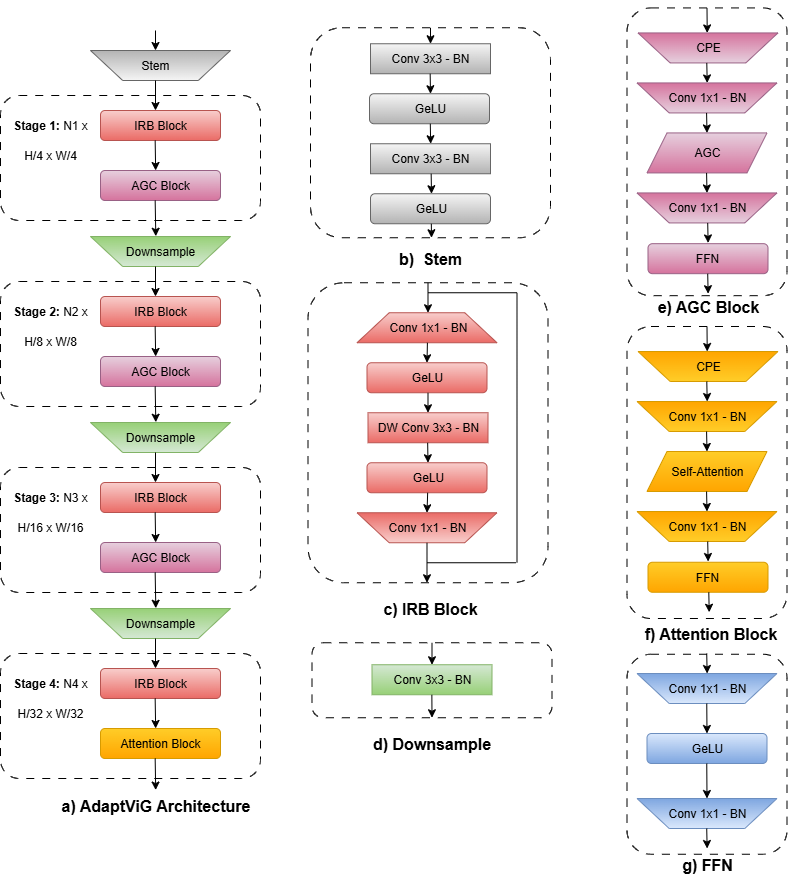}
    \caption{\textbf{The AdaptViG Architecture.} 
    \textbf{(a)} The overall 4-stage hierarchical architecture. Stages 1-3 use our AGC Block, while Stage 4 uses our attention block. 
    \textbf{(b)} The initial convolutional stem. 
    \textbf{(c)} The Inverted Residual Block (IRB) for local feature processing. 
    \textbf{(d)} The downsample block used between stages. 
    \textbf{(e)} The full AGC Block, which contains our AGC mixer and an FFN. 
    \textbf{(f)} The Attention Block, which replaces the AGC mixer with the self-attention mechanism. 
    \textbf{(g)} The Feed-Forward Network (FFN) used in both block types.}
    \label{fig:overall_arch}
    \vspace{-5mm}
\end{figure}

\vspace{-1mm}
\subsection{AGC and Attention Blocks}
\label{ssec:graph_block}
The global processing in AdaptViG is performed by two specialized modules: the \textbf{AGC Block} and the \textbf{Attention Block}. The complete designs for these are shown in Figure~\ref{fig:overall_arch}(e) and (f).

As depicted in Figure~\ref{fig:overall_arch}(e), the \textbf{AGC Block} first enriches features with positional information using a Conditional Positional Encoding (CPE)~\cite{CPE} layer. The features are then passed through a 1$\times$1 convolution before being processed by our proposed Adaptive Graph Convolution (AGC) mechanism. The \textbf{Attention Block}, shown in Figure~\ref{fig:overall_arch}(f), follows the exact same structure, but replaces the core AGC mechanism with a global self-attention mechanism. This makes it ideal for the final, low-resolution stage where maximal feature aggregation is desired. The output is then passed through a final Feed-Forward Network (FFN), shown in Figure~\ref{fig:overall_arch}(g). This design effectively combines the global information from the graph or attention operation with powerful feature transformations from the FFN.

\begin{table*}[ht]
\fontsize{10}{11}\selectfont
\def\arraystretch{1.0}
\caption{\textbf{Classification results on ImageNet-1k} for AdaptViG and other state-of-the-art models. Bold entries indicate results obtained for AdaptViG. The accuracy results for AdaptViG models are averaged over four runs and show the mean $\pm$ standard deviation.}
\centering
\begin{tabular}[t]{|c|c|c|c|c|c|c|}
\hline
{\textbf{Model}} & {\textbf{Type}} & {\textbf{Parameters (M)}} & {\textbf{GMACs}} & {\textbf{Epochs}} & {\textbf{Top-1 Acc (\%)}}       \\ \hline

ResNet18 \cite{Resnet} & CNN & 11.7 & 1.8  & 300 & 69.7 \\ \hline
ResNet50 \cite{Resnet}                   & CNN         & 25.6    & 4.1             & 300 & 80.4      \\ \hline
ConvNext-T \cite{liu2022convnet}                 & CNN         & 28.6    & 7.4  &     300 & 82.7      \\ \hline
 {HorNet-T \cite{HorNet}} &  {CNN} &  {22.0} &  {4.0} &  {300} &  {82.8} \\ \hline
 {HorNet-S \cite{HorNet}} &  {CNN} &  {50.0} &  {8.8}  &  {300} &  {83.8} \\ \hline
 {MogaNet-S \cite{MogaNet}} &  {CNN} &  {25.3} &  {5.0}  &  {300} &  {83.4} \\ \hline
 {MambaOut-Tiny \cite{MambaOut}} &  {CNN} &  {27.0} &  {4.5}  &  {300} &  {82.7} \\ \hline
EfficientFormer-L1 \cite{EfficientFormer}         & CNN-ViT     & 12.3    & 1.3  &   300 & 79.2      \\ \hline
EfficientFormer-L3 \cite{EfficientFormer}           & CNN-ViT     & 31.3    & 3.9  &   300 & 82.4      \\ \hline
EfficientFormer-L7 \cite{EfficientFormer}           & CNN-ViT     & 82.1    & 10.2 &    300 & 83.3    \\ \hline
LeViT-192 \cite{graham2021levit}          & CNN-ViT     & 10.9    & 0.7           & 1000 & 80.0      \\ \hline
LeViT-384 \cite{graham2021levit}                   & CNN-ViT     & 39.1    & 2.4            & 1000 & 82.6      \\ \hline
 {MaxViT-T \cite{MaxViT}} &  {CNN-ViT} &  {31.0} &  {5.6}  &  {300} &  {83.6} \\ \hline
PVT-Small \cite{wang2021pyramid}           & ViT         & 24.5        & 3.8     & 300 & 79.8 \\ \hline
PVT-Large \cite{wang2021pyramid}           & ViT         & 61.4        & 9.8           & 300 & 81.7 \\ \hline
Swin-T \cite{liu2021swin}                   & ViT   & 29.0    & 4.5       & 300 & 81.3      \\ \hline
Swin-S \cite{liu2021swin}                   & ViT   & 50.0    & 8.7          & 300 & 83.0      \\ \hline
 {CrossViT-15 \cite{CrossViT}} &  {ViT} &  {27.4} &  {5.8} &  {300} &  {81.5} \\ \hline
 {CrossViT-18 \cite{CrossViT}} &  {ViT} &  {43.3} &  {9.0} &  {300} &  {82.5} \\ \hline
 {Vim-Ti \cite{Vim}} &  {SSM} &  {7.0} &  {-}  &  {300} &  {76.1} \\ \hline
 {Vim-S \cite{Vim}} &  {SSM} &  {26.0} &  {-}  &  {300} &  {80.3} \\ \hline
 {Vim-B \cite{Vim}} &  {SSM} &  {98.0} &  {-}  &  {300} &  {81.9} \\ \hline
 {VMamba-T \cite{VMamba}} &  {SSM} &  {30.0} &  {4.9}  &  {300} &  {82.6} \\ \hline
 {VMamba-S \cite{VMamba}} &  {SSM} &  {50.0} &  {8.7}  &  {300} &  {83.6} \\ \hline
 {VMamba-B \cite{VMamba}} &  {SSM} &  {89.0} &  {15.4}  &  {300} &  {83.9} \\ \hline
PoolFormer-s12 \cite{MetaFormer}         & Pool        & 12.0    & 2.0         & 300 & 77.2      \\ \hline
PoolFormer-s24 \cite{MetaFormer}               & Pool        & 21.0    & 3.6         & 300 & 80.3      \\ \hline
PoolFormer-s36 \cite{MetaFormer}               & Pool        & 31.0    & 5.2            & 300 & 81.4      \\ \hline
PViHGNN-Ti  \cite{ViHGNN}     & GNN & 12.3    & 2.3  & 300 & 78.9      \\ \hline
PViHGNN-S   \cite{ViHGNN}    & GNN & 28.5    & 6.3 & 300 & 82.5      \\ \hline
PViHGNN-B   \cite{ViHGNN}    & GNN & 94.4 & 18.1      & 300 & 83.9      \\ \hline
ViG-S \cite{Vision_GNN} & GNN & 22.7 & 4.5 & 300 & 80.4 \\ \hline
ViG-B \cite{Vision_GNN} & GNN & 86.8 & 17.7  & 300 & 82.3 \\ \hline
PViG-Ti \cite{Vision_GNN}       & GNN & 10.7    & 1.7  & 300 & 78.2      \\ \hline
PViG-S \cite{Vision_GNN}       & GNN & 27.3    & 4.6  & 300 & 82.1      \\ \hline
PViG-B \cite{Vision_GNN}       & GNN & 92.6 & 16.8     & 300 & 83.7      \\ \hline
 {PVG-S \cite{PVG}} &  {GNN} &  {22.0} &  {5.0} &  {300} &  {83.0} \\ \hline
 {PVG-M \cite{PVG}} &  {GNN} &  {42.0} &  {8.9}  &  {300} &  {83.7} \\ \hline
 {PVG-B \cite{PVG}} &  {GNN} &  {79.0} &  {16.9} &  {300} &  {84.2} \\ \hline
{MobileViG-S} \cite{MobileViG}   & {CNN-GNN}     & {7.2}     & {1.0}  &  {300} & {78.2}      \\ \hline
{MobileViG-M} \cite{MobileViG}     & {CNN-GNN}     & {14.0}        & {1.5}   & {300} & {80.6}      \\ \hline
{MobileViG-B} \cite{MobileViG}   & {CNN-GNN}     & {26.7}        & {2.8}    & {300} & {82.6}      \\ \hline
\rowcolor {Gray}
\textbf{AdaptViG-S (Ours)}   & \textbf{CNN-GNN}     & \textbf{8.6}     & \textbf{1.0}    & \textbf{300} & \textbf{79.6 $\pm$ 0.1}      \\ \hline
\rowcolor {Gray}
\textbf{AdaptViG-M (Ours)}     & \textbf{CNN-GNN}     & \textbf{17.9}      & \textbf{2.9}    & \textbf{300} & \textbf{82.6 $\pm$ 0.1}      \\ \hline
\rowcolor {Gray}
\textbf{AdaptViG-B (Ours)}     & \textbf{CNN-GNN}     & \textbf{26.8}      & \textbf{3.8}    & \textbf{300} & \textbf{83.3 $\pm$ 0.1}      \\ \hline
\end{tabular}
\label{Classification_Results}
\end{table*}

\vspace{-1mm}
\subsection{AdaptViG Architecture}
\label{ssec:architecture}
The overall AdaptViG architecture, shown in Figure~\ref{fig:overall_arch}(a), is a hierarchical hybrid design composed of a convolutional stem and four subsequent stages that progressively reduce spatial resolution while increasing channel width.

The \textbf{Stem} (Figure~\ref{fig:overall_arch}(b)) uses two 3$\times$3 convolutions, each with a stride of 2, to quickly downsample the input image by a factor of 4 and extract initial low-level features.

Each of the four \textbf{Stages} contains a sequence of blocks for local and global processing. For local processing, we use efficient \textbf{Inverted Residual} blocks from MobileNetV2~\cite{MobileNetv2}, as shown in Figure~\ref{fig:overall_arch}(c). For global processing, a key element of our design is the hybrid strategy: Stages 1-3 use our efficient \textbf{AGC Block} (Fig.~\ref{fig:overall_arch}e), while the final, lowest-resolution stage uses the more expensive \textbf{Attention Block} (Fig.~\ref{fig:overall_arch}f) to maximize feature aggregation when the computational cost is lowest. Between each stage, a \textbf{Downsample} block (Figure~\ref{fig:overall_arch}(d)), consisting of a strided 3$\times$3 convolution, is used to halve the feature map resolution and expand the channel dimension. The architecture terminates with a classification head composed of adaptive average pooling and fully connected layers. The detailed network configurations are available in the Supplementary Section \ref{Sec:Suppl_configuration}.


\begin{table*}[ht]
\small
\def\arraystretch{1.0}
\caption{\textbf{Object detection, instance segmentation, and semantic segmentation results} of AdaptViG and other backbones on MS COCO 2017 and ADE20K. Bold entries indicate results obtained using AdaptViG. A (-) denotes a model that did not report these results.}
\centering
\begin{tabular}[t]{|c|c|c|c|c|c|c|c|c|c|c|}
\hline
{\textbf{Backbone}} & {\textbf{Parameters (M)}} & \textbf{$AP^{box}$} & \textbf{$AP^{box}_{50}$} & \textbf{$AP^{box}_{75}$} & \textbf{$AP^{mask}$} & \textbf{$AP^{mask}_{50}$} & \textbf{$AP^{mask}_{75}$} & \textbf{$mIoU$} \\ \hline
         
ResNet18 \cite{Resnet}     & 11.7  & 34.0 & 54.0 & 36.7 & 31.2 & 51.0 & 32.7 & 32.9    \\ \hline
EfficientFormer-L1 \cite{EfficientFormer}  & 12.3 & 37.9 & 60.3 & 41.0 & 35.4 & 57.3 & 37.3 & 38.9        \\ \hline
PoolFormer-S12 \cite{MetaFormer}  & 12.0  & 37.3 & 59.0 & 40.1 & 34.6 & 55.8 & 36.9  & 37.2        \\ \hline
FastViT-SA12 \cite{FastViT}  & 10.9 & 38.9 & 60.5 & 42.2 & 35.9 & 57.6 & 38.1 & 38.0        \\ \hline
{MobileViG-M} \cite{MobileViG}   & {14.0} & {41.3} & {62.8} & {45.1} & {38.1} & {60.1} & {40.8} & -   \\ \hline
PoolFormer-S24 \cite{MetaFormer}         & 21.0 & 40.1 & 62.2 & 43.4 & 37.0 & 59.1 & 39.6 & 40.3   \\ \hline
\rowcolor{Gray}
\textbf{AdaptViG-M (Ours)} & \textbf{17.9} & \textbf{44.8} & \textbf{67.0} & \textbf{49.4} & \textbf{41.1} & \textbf{64.3} & \textbf{44.2}  & \textbf{45.8}   \\ \hline
ResNet50 \cite{Resnet}         & 25.5 & 38.0 & 58.6 & 41.4 & 34.4 & 55.1 & 36.7 & 36.7    \\ \hline
EfficientFormer-L3 \cite{EfficientFormer}   & 31.3 & 41.4 & 63.9 & 44.7 & 38.1 & 61.0 & 40.4 & 43.5    \\ \hline
EfficientFormer-L7 \cite{EfficientFormer}   & 82.1 &  42.6 & 65.1 & 46.1 & 39.0 & 62.2 & 41.7 & 45.1        \\ \hline
FastViT-SA36 \cite{FastViT}               & 30.4 & 43.8 & 65.1 & 47.9 & 39.4 & 62.0 & 42.3 & 42.9 \\ \hline
Pyramid ViG-S \cite{Vision_GNN}      & 27.3 & 42.6 & 65.2 & 46.0 & 39.4 & 62.4 & 41.6 & -   \\ \hline
Pyramid ViHGNN-S \cite{ViHGNN}          & 28.5 & 43.1 & 66.0 & 46.5 & 39.6 & 63.0 & 42.3 & - \\ \hline
PVT-Small \cite{wang2021pyramid}     & 24.5 & 40.4 & 62.9 & 43.8 & 37.8 & 60.1 & 40.3  & 39.8  \\ \hline
PVT-Large \cite{wang2021pyramid}     & 61.4 & 42.9 & 65.0 & 46.6 & 39.5 & 61.9 & 42.5  & 42.1  \\ \hline
{MobileViG-B} \cite{MobileViG}  & {26.7} & {42.0} & {64.3} & {46.0} & {38.9} & {61.4} & {41.6} & -   \\ \hline
\rowcolor{Gray}
\textbf{AdaptViG-B (Ours)} & \textbf{26.8} & \textbf{46.0} & \textbf{68.0} & \textbf{50.7} & \textbf{42.1} & \textbf{65.2} & \textbf{45.1}  & \textbf{47.4}   \\ \hline
\end{tabular}
\label{Object_Detection_Segmentation_Results}
\vspace{-4mm}
\end{table*}

\vspace{-2mm}

\section{Experimental Results}
\label{sec:experiments}

\vspace{-2mm}

We compare AdaptViG with competing CNN, ViT, and Vision GNN architectures on the tasks of image classification, object detection, instance segmentation, and semantic segmentation to demonstrate its superior performance. For additional results on the CIFAR \cite{Cifar} and MedMNIST (OrganSMNIST and DermaMNIST)\cite{yang2023medmnist} benchmarks, please refer to the Supplementary Materials Section \ref{Sec:Suppl_Experimental}.

\vspace{-2mm}

\subsection{Image Classification}

\vspace{-2mm}

All AdaptViG models are trained from scratch on the ImageNet-1K dataset~\cite{imagenet1k} using PyTorch~\cite{paszke2019pytorch} and the Timm library~\cite{timm}. We train for 300 epochs using a batch size of 1024, the AdamW optimizer~\cite{Adam, AdamW}, and a learning rate of 2e$^{-3}$ with cosine annealing schedule. We use a standard training resolution of $224 \times 224$.

As shown in Table~\ref{Classification_Results}, AdaptViG consistently provides better performance than prior Vision GNNs. Our smallest model, AdaptViG-S, achieves 79.6\% Top-1 accuracy, outperforming PViG-Ti~\cite{Vision_GNN} and MobileViG-S~\cite{MobileViG} by 1.4\% with much fewer GMACs. Our mid-sized model, AdaptViG-M, reaches 82.6\% accuracy, matching MobileViG-B~\cite{MobileViG} while using 33\% fewer parameters. Our largest model, AdaptViG-B, achieves 83.3\% accuracy with only 26.8M parameters, matching the much larger EfficientFormer-L7~\cite{EfficientFormer} (83.3\%) while using 67\% fewer parameters and 63\% fewer GMACs. When compared to other architectures, AdaptViG-B also outperforms the larger Swin-S~\cite{liu2021swin} (83.0\%) and ConvNext-T~\cite{liu2022convnet} (82.7\%).

\vspace{-2mm}

\subsection{Object Detection and Instance Segmentation}

\vspace{-2mm}

To evaluate generalization to downstream tasks, we use AdaptViG as a backbone in the Mask-RCNN framework \cite{mask_r_cnn} for object detection and instance segmentation tasks on the MS COCO 2017 dataset \cite{coco}. The dataset contains training and validations sets of 118K and 5K images, respectively. We implement the backbone using PyTorch 1.12.1 \cite{paszke2019pytorch} and Timm library \cite{timm}. The model is initialized with ImageNet-1k pretrained weights from 300 epochs of training. We use the AdamW \cite{Adam, AdamW} optimizer with an initial learning rate of 2e$^{-4}$ and train the model for 12 epochs with a standard resolution (1333 $\times$ 800) following the process of prior work \cite{li2022next, MobileViG, EfficientFormer, EfficientFormerv2}.

\begin{figure*}[t!]
    \centering
    \includegraphics[width=\textwidth]{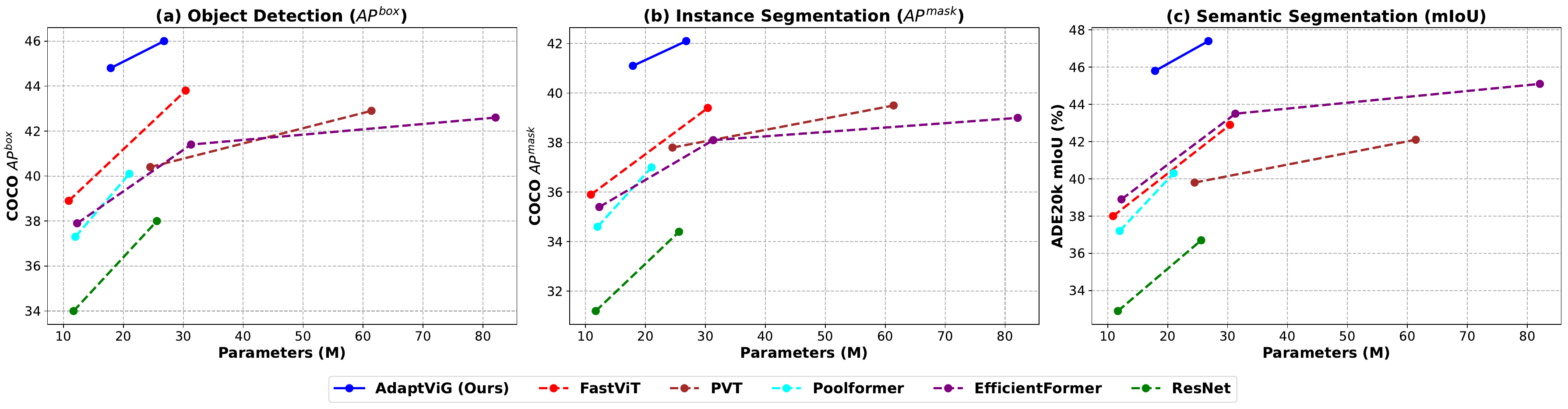}
    \caption{\textbf{Downstream task performance vs. model size.} AdaptViG establishes a new state-of-the-art Pareto frontier on all three downstream tasks, outperforming competing backbones. (a) Object detection performance ($AP^{box}$) on MS-COCO. (b) Instance segmentation performance ($AP^{mask}$) on MS-COCO. (c) Semantic segmentation performance (mIoU) on ADE20K.}
    \label{fig:pareto_downstream}
    \vspace{-4mm}
\end{figure*}

The results in Table~\ref{Object_Detection_Segmentation_Results} demonstrate AdaptViG's strong performance. AdaptViG-M achieves 44.8 $AP^{box}$ and 41.1 $AP^{mask}$, significantly outperforming other models in its size class like PoolFormer-S24~\cite{MetaFormer} by +4.7 $AP^{box}$ and +4.1 $AP^{mask}$. Our AdaptViG-B model achieves 46.0 $AP^{box}$ and 42.1 $AP^{mask}$, surpassing larger and more computationally expensive models like EfficientFormer-L3~\cite{EfficientFormer} by +4.6 $AP^{box}$ and +4.0 $AP^{mask}$. The performance of AdaptViG across these tasks, visualized in Figure~\ref{fig:pareto_downstream}(a-b), underscores the strong representation capabilities of our proposed architecture.

\vspace{-2mm}

\subsection{Semantic Segmentation}

\vspace{-2mm}

We further validate our models on semantic segmentation using the ADE20K dataset~\cite{ADE20K}, which contains 20K training and 2K validation images across 150 semantic categories. Following the methodologies of prior work~\cite{MetaFormer, EfficientFormer, EfficientFormerv2, FastViT}, we build AdaptViG with Semantic FPN~\cite{kirillov2019panoptic} as the segmentation decoder. The AdaptViG backbones are initialized with their ImageNet-1K pre-trained weights and trained for 40K iterations on 8 NVIDIA RTX 6000 Ada generation GPUs. We use the AdamW optimizer with a learning rate of $2 \times 10^{-4}$ and a polynomial decay schedule with a power of 0.9. All models are trained with a resolution of $512 \times 512$~\cite{EfficientFormer, EfficientFormerv2}.

As shown in Table~\ref{Object_Detection_Segmentation_Results} and Figure~\ref{fig:pareto_downstream}(c), our models establish a new state-of-the-art Pareto frontier. AdaptViG-M achieves 45.8 mIoU, which is +2.3 mIoU higher than the larger EfficientFormer-L3~\cite{EfficientFormer}. Furthermore, our AdaptViG-B model obtains 47.4 mIoU, outperforming the much larger EfficientFormer-L7~\cite{EfficientFormer} by +2.3 mIoU with less than one-third of the parameters (26.8M vs. 82.1M). These results confirm that the features learned by AdaptViG are highly effective for dense prediction tasks.

\vspace{-1mm}

\subsection{Ablation Studies}
\label{sec:ablation}

To validate our design choices, we conduct ablation studies on the ImageNet-1K \cite{imagenet1k} dataset using our AdaptViG-M model as the baseline. Table~\ref{tab:ablation_study} reports the results of ablating key components of our architecture, including the gating mechanism and the hybrid graph mixer strategy.

\vspace{-2mm}

\begin{table}[ht]
    \vspace{-1mm}
    \centering
    \caption{\textbf{Ablation study on AdaptViG-M} components, trained on ImageNet-1K for 300 epochs.}
    \def\arraystretch{0.95}
    \resizebox{\columnwidth}{!}{%
        \begin{tabular}{lcc}
        \toprule
        \textbf{Configuration} & \textbf{Params (M)} & \textbf{Top-1 Acc (\%)} \\
        \midrule
        \rowcolor{Gray}
        \textbf{AdaptViG-M} & \textbf{17.9} & \textbf{82.6} \\
        \midrule
        \textit{Graph Mixer Variants:} & & \\
        Static Graph Only (Config 1) & 17.3 & 81.5 \\
        Gating in All Stages (Config 2) & 17.3 & 82.5 \\
        Attention in All Stages (Config 3) & 18.3 & 81.2 \\
        \bottomrule
        \end{tabular}%
    }
    \label{tab:ablation_study}
    \vspace{-6mm}
\end{table}

\vspace{-2mm}

\paragraph{Effectiveness of Gating.}
As shown in Table~\ref{tab:ablation_study}, the dynamic nature of our graph construction is critical for performance. Removing our gating mechanism and relying on a purely static graph (Config 1) results in a significant 1.1\% drop in Top-1 accuracy.

This performance degradation highlights a limitation of static topologies, which are more susceptible to \textbf{over-squashing} \cite{alon2020bottleneck}. Oversquashing occurs when information from a large receptive field must be compressed into a fixed-size vector at each step, causing information from distant nodes to be lost. This is a known bottleneck in GNNs and static graph methods \cite{MobileViG, LogViG}. Our dynamic gating helps alleviate this issue by creating efficient, content-aware shortcuts, confirming the importance of our approach.

\vspace{-4mm}

\paragraph{Impact of Hybrid Strategy.}
Our hybrid approach of using efficient gating in early stages and powerful global attention in the final stage proves to be highly effective. Using our efficient gating mechanism in all stages (Config 2) performs well but results in a 0.1\% accuracy drop compared to the full model, indicating that the final global attention block provides a final feature aggregation step that boosts performance. Conversely, using the expensive global attention block in all stages (Config 3) not only increases the parameter count but also degrades accuracy by a substantial 1.4\%. This poor performance is likely due to the computational cost of applying dense attention on high-resolution feature maps. This ablation validates our hybrid design of using the most suitable mixer for each stage of the network.

Additional ablations, including an analysis of the static connection distance $K$ and computational complexity of the different graph construction methods are detailed in the Supplementary Materials Section \ref{app:Additional_Ablations}.


\vspace{-2mm}

\section{Conclusion}
\label{sec:conclusion}
\vspace{-2mm}
In this paper, we have introduced AdaptViG, a novel hybrid Vision GNN architecture featuring the novel Adaptive Graph Convolution (AGC) mechanism. AGC creates an efficient static scaffold and dynamically weighs long-range connections based on feature similarity using an Exponential Decay Gating mechanism. AdaptViG leverages AGC in its early, high-resolution stages for efficiency and transitions to a global attention block in its final stage to maximize feature aggregation at low-resolution. This hybrid strategy resolves the critical trade-off between adaptability and efficiency found in prior Vision GNNs.

Our extensive experiments demonstrate that AdaptViG establishes a new state-of-the-art trade-off between accuracy and efficiency on ImageNet-1K classification and key downstream tasks. For instance, our AdaptViG-B achieves 83.3\% top-1 accuracy, outperforming the much larger ViG-B by 1.0\% while using 69\% fewer parameters. On downstream tasks, AdaptViG-B obtains 47.4 mIoU on ADE20K and 46.0 $AP^{box}$ on MS-COCO, surpassing comparable models. These results validate AdaptViG as an efficient backbone, demonstrating that a well-designed hybrid GNN can outperform leading CNN and ViT-based models.

{
    \small
    \bibliographystyle{ieeenat_fullname}
    \bibliography{main}
}

\clearpage
\setcounter{page}{1}
\maketitlesupplementary
\appendix
\section{Theoretical Basis of Adaptive Graph Convolution}
\label{app:theory}

\subsection{Foundations of Graph-Based Information Aggregation}
\label{app:foundations}

In vision models, a feature map can be viewed as a grid graph where each pixel (or patch embedding) is a node. A graph convolution operation aims to update a node's feature vector by aggregating information from its neighbors. Let $X \in \mathbb{R}^{B \times C \times H \times W}$ be an input feature map. For a node $v_p$ at position $p$ with feature vector $x_p \in \mathbb{R}^C$, a general graph update rule can be written as:
\begin{equation}
    x'_p = \phi \left( x_p, \bigoplus_{q \in \mathcal{N}(p)} \psi(x_p, x_q) \right)
\end{equation}
where $\mathcal{N}(p)$ is the set of neighbors of node $p$, $\psi$ is a message function computing the relationship between nodes, $\bigoplus$ is an aggregation function (e.g., sum, mean, max), and $\phi$ is an update function (e.g., an MLP).

Our AGC (Adaptive Graph Convolution) module implements a specific form of this update. We define the aggregated information from all neighbors as a single "max-relative feature" tensor $X_j \in \mathbb{R}^{B \times C \times H \times W}$. The message function $\psi$ is the feature difference $(x_q - x_p)$, and the aggregation $\bigoplus$ is the element-wise maximum. The final update $\phi$ is a concatenation followed by a convolutional block.

\subsection{The Static Scaffold: An Efficient Structural Prior}
\label{app:scaffold}

The foundation of our AGC module is a highly efficient, predefined graph structure we term the \textbf{static scaffold}. This scaffold defines the neighborhood set $\mathcal{N}(p)$ for every node $p$ and is designed to capture a rich set of spatial relationships without relying on computationally expensive search algorithms like KNN. The scaffold is composed of two distinct types of connections, ensuring a comprehensive receptive field. The aggregated feature map for the scaffold, $X_{j, \text{static}}$, is initialized as a zero tensor.

\paragraph{1. Fixed Local Connections} To ensure a strong inductive bias for local feature interactions, we establish connections to neighbors at a fixed distance, $K$. This is similar to a dilated convolution and is crucial for capturing fine-grained details. Let $\text{roll}(X, s, a)$ be an operator that cyclically shifts tensor $X$ by $s$ positions along axis $a$. The update rule uses one shift per spatial axis:
\begin{align}
    X_{j} &\leftarrow \max(X_{j}, \text{roll}(X, -K, H) - X) \\
    X_{j} &\leftarrow \max(X_{j}, \text{roll}(X, -K, W) - X)
\end{align}
where $H$ and $W$ correspond to height and width axes where the roll operations occur.

\paragraph{2. Logarithmic Long-Range Connections} To model the global context efficiently, we form connections to neighbors at exponentially increasing distances. This allows information to propagate across the entire feature map in a logarithmic number of steps, rather than linear. For each spatial dimension, we iterate and update $X_j$:

\footnotesize{
\begin{align}
    \forall i \in [1, \lfloor\log_2 H\rfloor]: X_{j} &\leftarrow \max(X_{j}, \text{roll}(X, -2^i, 2) - X) \\
    \forall i \in [1, \lfloor\log_2 W\rfloor]: X_{j} &\leftarrow \max(X_{j}, \text{roll}(X, -2^i, 3) - X)
\end{align}}
\normalsize
The combination of these two strategies creates the static scaffold, providing a robust structural prior that captures both short- and long-range dependencies with minimal computational overhead.

\subsection{Exponential Decay Gating for Content-Awareness}
\label{app:gating}

While the static scaffold is efficient, it is not content-aware. To address this, we introduce a \textit{dynamic gating} mechanism that weighs connections based on semantic similarity. The core of this mechanism is the \textbf{Exponential Decay Gate}, a function designed for numerical stability.

For a given patch, $x_p \in \mathbb{R}^C$ and a potential neighbor $x_n$, we define a gate value $g_{pn} \in (0, 1]$ that modulates their connection.

\paragraph{Definition A.1: Feature Distance.}
We first compute the dissimilarity between the two patches using the L1 norm, which is computationally efficient and robust to outliers.
\begin{equation}
    d(x_p, x_n) = ||x_p - x_n||_1 = \sum_{c=1}^{C} |x_{p,c} - x_{n,c}|
\end{equation}

\paragraph{Definition A.2: Exponential Decay Gate.}
The distance $d(x_p, x_n)$ is then transformed into a gating value using an exponential decay function.
\begin{equation}
    g(x_p, x_n) = \exp\left(-\frac{d(x_p, x_n)}{T}\right)
    \label{eq:exp_decay}
\end{equation}
Here, $T$ is a learnable scalar parameter (\textbf{temperature}).

\paragraph{Rationale and Numerical Stability.}
A key motivation for this gating function is to avoid numerical instability, which often leads to `NaN` losses during training.
\begin{itemize}
    \item \textbf{No Division by Features:} The function in Equation \ref{eq:exp_decay} does not divide by any value derived from the input features. This is critical as it ensures there will be no divide by zero cases even if the input has zero variance (e.g., a uniform patch in an image).
    \item \textbf{Bounded Output:} The output of the exponential gate is naturally bounded. Since the L1 distance is always non-negative ($d \ge 0$) and we enforce $T > 0$, the argument to `exp` is always non-positive $(-\frac{d}{T} \le 0)$. This guarantees the output is bounded within the range $(0, 1]$ and cannot explode to infinity.
    \item \textbf{Learnable Sensitivity:} The temperature $T$ provides crucial flexibility. We constrain it to be positive by using $|T| + \epsilon$ in the implementation. The model learns the optimal sensitivity for each layer.
\end{itemize}


\subsection{Gradient Flow Analysis}
\label{app:loss}

To understand how the network learns to form meaningful connections, we discuss the gradient flow from the loss function $\mathcal{L}$ back to the parameters of the gating mechanism, specifically the temperature $T$.

Let us consider the contribution of a single gated connection to the aggregated feature $X_j$. Let the message from a neighbor be $M_{pn} = x_n - x_p$. The gated message is $M'_{pn} = g_{pn} \odot M_{pn}$ where $\odot$ is the element-wise product. The update to $X_j$ at position $p$ involves this term. For simplicity, let's assume the loss $\mathcal{L}$ is a function of $X_j$. The gradient of the loss with respect to the gate value $g_{pn}$ is given by the chain rule:
\begin{equation}
    \frac{\partial \mathcal{L}}{\partial g_{pn}} = \frac{\partial \mathcal{L}}{\partial X_j(p)} \frac{\partial X_j(p)}{\partial M'_{pn}} \frac{\partial M'_{pn}}{\partial g_{pn}}
\end{equation}
Since $\frac{\partial M'_{pn}}{\partial g_{pn}} = M_{pn} = (x_n - x_p)$, this becomes:
\begin{equation}
    \frac{\partial \mathcal{L}}{\partial g_{pn}} = \nabla_{X_j(p)}\mathcal{L} \cdot (x_n - x_p)
\end{equation}
This shows that the gradient for a gate depends on how much the feature difference $(x_n - x_p)$ aligns with the overall gradient of the loss. If incorporating the neighbor's information helps reduce the loss, the gradient will push the gate value $g_{pn}$ higher, strengthening the connection.

Now, let's analyze the gradient for the temperature $T$, which controls the gate's sensitivity. Using the chain rule again:
\begin{equation}
    \frac{\partial \mathcal{L}}{\partial T} = \sum_{p,n} \frac{\partial \mathcal{L}}{\partial g_{pn}} \frac{\partial g_{pn}}{\partial T}
    \label{eq:grad_T_1}
\end{equation}
The derivative of the gate function with respect to $T$ is:
\begin{equation}
\footnotesize
    \frac{\partial g_{pn}}{\partial T} = \frac{\partial}{\partial T} \exp\left(\frac{-d_{pn}}{T}\right) = \exp\left(\frac{-d_{pn}}{T}\right) \cdot \frac{d_{pn}}{T^2} = g_{pn} \frac{d_{pn}}{T^2}
\end{equation}
Substituting this back in Equation \ref{eq:grad_T_1}, we get:
\begin{equation}
    \frac{\partial \mathcal{L}}{\partial T} = \sum_{p,n} \left( \nabla_{X_j(p)}\mathcal{L} \cdot (x_n - x_p) \right) \left( g_{pn} \frac{d_{pn}}{T^2} \right)
    \label{eq:grad_T}
\end{equation}
The update rule for $T$ during optimization will be $T \leftarrow T - \eta \frac{\partial \mathcal{L}}{\partial T}$. \Cref{eq:grad_T} reveals that the learning signal for $T$ is proportional to the distance $d_{pn}$. This means the model learns a single temperature $T$ that best balances the trade-off across all connections: it will be pushed to a value that appropriately suppresses connections with large, unhelpful distances while preserving connections with small, useful distances, all in service of minimizing the global loss $\mathcal{L}$. This provides a strong theoretical justification for the mechanism's ability to learn meaningful graph structures.


\section{Additional Ablation Studies}
\label{app:Additional_Ablations}

We conduct ablation studies on ImageNet-1K~\cite{imagenet1k} to validate the key design choices of our AdaptViG architecture. Unless otherwise specified, the ablations are performed on the AdaptViG-B model.

\subsection{Static Connection Distance (K).}

The distance $K$ for the guaranteed local connections in the static scaffold of our AGC blocks can affect performance. We test various configurations for the $K$ values used in Stages 1, 2, and 3. As shown in Table~\ref{tab:ablation_k_distance}, we find that using $K$ values of [8, 4, 2] for these three stages, respectively, yields the best performance. Increasing the initial hop distance ($K$ = [16, 8, 4]) results in a 0.1\% accuracy drop, while a different pattern ($K$ = [9, 6, 3]) leads to a 0.1\% drop, validating our chosen configuration.

\begin{table}[ht]
\def\arraystretch{1.2}
\caption{\textbf{Ablation on the static connection distance (K)} for the AGC blocks in AdaptViG-B. The Top-1 accuracy results are averaged over two experiments.}
\centering
\begin{tabular}[t]{|c|c|c|}
\hline
\textbf{K (for AGC Stages 1-3)} & \textbf{Params (M)} & \textbf{Top-1 (\%)} \\ \hline
K = 16, 8, 4  & 26.8    & 83.2 \\
K = 9, 6, 3   & 26.8    & 83.2 \\
\rowcolor{Gray}
\textbf{K = 8, 4, 2}  & \textbf{26.8}  & \textbf{83.3}  \\ \hline
\end{tabular}
\label{tab:ablation_k_distance}
\end{table}

\subsection{Computational Complexity}

The computational complexity of the graph construction step is a critical factor in the efficiency of Vision GNNs. Here, we analyze the per-node complexity for different methods. Let $W$ and $H$ be the width and height of the feature map, $K$ be the number of nearest neighbors for KNN, and $N$ is the number of hops selected in SVGA and DAGC.

\begin{enumerate}
    \item \textbf{KNN} \cite{Vision_GNN, KNN}: O($W \times H$). For each node, KNN must compare against all other nodes in the feature map, making it computationally prohibitive.
    
    \item \textbf{DAGC} \cite{GreedyViG}: O($\frac{W + H}{N}$). This method connects nodes at linear intervals with a stride of N, making its complexity dependent on the feature map's dimensions divided by the hop distance.

    \item \textbf{AGC (Ours)}: O($\log W + \log H$). Our method connects nodes at logarithmic intervals, making it highly efficient and scalable while still capturing global, content-aware relationships.
    
    \item \textbf{SVGA} \cite{MobileViG}: O(1). As a static method, the graph is fixed, and the connections are retrieved via efficient shift operations with constant time complexity per node.
\end{enumerate}

Our AGC is marginally more expensive than a purely static method like SVGA, but is more efficient than both KNN and DAGC. To demonstrate this, we measure the graph construction time of a single forward pass on an Nvidia RTX A6000. We measure an average time of 0.048 ms for our AGC, which is 20\% faster than DAGC (0.06 ms) and significantly faster than KNN (0.38 ms), while being nearly on par with the static SVGA (0.04 ms).

This efficiency translates into better end-to-end model latency per batch, as shown in Table~\ref{tab:latency_graph_construction}. Our \textbf{AdaptViG-S w/ AGC} is significantly faster than a comparable model using KNN and is also more accurate and faster than PViG-Ti using KNN. This highlights the effectiveness of our architecture and the efficiency of the AGC module.

\begin{table}[ht]
\footnotesize
\def\arraystretch{1.2}
\caption{\textbf{Graph construction impact on accuracy and latency.} The table shows that our AGC provides a better accuracy-latency trade-off compared to KNN, DAGC, and static SVGA methods for models of similar size.}
\centering
\begin{tabular}[t]{|c|c|c|c|}
\hline
\textbf{Model} & \textbf{Params} & \textbf{Latency} & \textbf{Acc (\%)} \\ \hline
MobileViG-S~\cite{MobileViG} w/ SVGA & 7.2 M  & 27.1 ms          & 78.2 \\
PViG-Ti~\cite{Vision_GNN} w/ KNN    & 10.7 M & 79.4 ms          & 78.2 \\
AdaptViG-S (Ours) w/ KNN        & 8.6 M  & 71.2 ms          & 78.9 \\
\rowcolor{Gray}
\textbf{AdaptViG-S (Ours) w/ AGC} & \textbf{8.6 M}  & \textbf{42.3 ms} & \textbf{79.6} \\ \hline
\end{tabular}
\label{tab:latency_graph_construction}
\end{table}

\subsection{Throughput Comparisons}
\label{app:throughput_comp}

\begin{figure}[htbp]
\centering
\includegraphics[width=0.9\columnwidth]{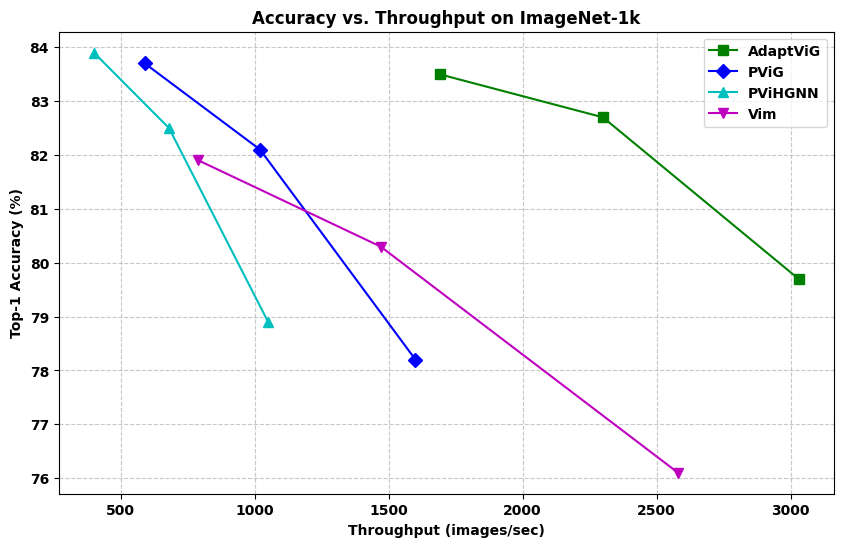}
\caption{\textbf{Accuracy vs. Throughput comparison on ImageNet-1k.} We plot Top-1 Accuracy against inference throughput (images/sec). AdaptViG clearly establishes a new state-of-the-art Pareto frontier, demonstrating significantly higher throughput and accuracy than competing models.}
\label{fig:throughput_viz}
\end{figure}

To validate the practical efficiency of our model, we measure the inference throughput. We compare our AdaptViG models against other Vision GNNs (PViG \cite{Vision_GNN}, PViHGNN \cite{ViHGNN}) and State Space Models (Vim) \cite{Vim} in Figure~\ref{fig:throughput_viz}.

As shown in Figure~\ref{fig:throughput_viz}, our AdaptViG models clearly establish a new state-of-the-art Pareto frontier, outperforming all competing architectures. Notably, our most efficient model, AdaptViG-S (79.6\%), achieves a throughput of over 3000 images/sec, which is more than twice as fast as Vim-S \cite{Vim} (80.3\%) and approximately three times as fast as PViG-S \cite{Vision_GNN} (82.1\%). This demonstrates the superior efficiency of our AdaptViG architecture.

\subsection{Ablation on Gating Temperature (T)}
\label{app:ablation_temp}

To analyze the impact of our learnable temperature parameter $T$, we conduct an ablation study presented in Table~\ref{tab:ablation_temp}. We compare our standard model, which learns $T$ per layer, against variants with a fixed, non-learnable $T$ set to different values. The results show that making $T$ learnable yields the best performance (83.3\%). While the model is reasonably robust to different fixed values, a very low temperature (e.g., $T=0.5$) is overly restrictive and hurts performance the most, while a higher temperature (e.g., $T=2.0$) can be too permissive. This validates our choice to make the temperature an adaptive, learnable parameter.

\begin{table}[ht]
\def\arraystretch{1.2}
\caption{{\textbf{Ablation on the temperature parameter (T)} in AdaptViG-B. Our learnable setting is optimal. The Top-1 accuracy results are averaged over two experiments.}}
\centering
\begin{tabular}[t]{|l|c|c|}
\hline
{\textbf{Configuration}} & {\textbf{Params (M)}} & {\textbf{Top-1 (\%)}} \\ \hline
{Fixed $T=0.5$} & {26.8} & {83.1} \\
{Fixed $T=1.0$} & {26.8} & {83.2} \\
{Fixed $T=2.0$} & {26.8} & {83.2} \\
\rowcolor{Gray}
{\textbf{Learnable T (Ours)}} & {\textbf{26.8}} & {\textbf{83.3}} \\ \hline
\end{tabular}
\label{tab:ablation_temp}
\end{table}

\textbf{Learned Temperature Ablation}. To analyze the behavior of our learnable temperature parameter, we inspect its final converged value for our AdaptViG-B model. The average learned value for each stage is reported in Table~\ref{tab:ablation_temp_learned}.

\begin{table}[ht]
\def\arraystretch{1.2}
\caption{\textbf{Average learned temperature (T) values} per stage for the \textbf{AdaptViG-B} model after 300 epochs of training on ImageNet-1K.}
\centering
\begin{tabular}[t]{|l|c|c|c|}
\hline
{\textbf{Model}} & {\textbf{Stage 1}} & {\textbf{Stage 2}} & {\textbf{Stage 3}} \\ \hline
{AdaptViG-B} & {3.93} & {0.68} & {1.11} \\ \hline
\end{tabular}
\label{tab:ablation_temp_learned}
\end{table}

The results in Table~\ref{tab:ablation_temp_learned} reveal a clear trend across the network's depth. In \textbf{Stage 1}, the temperature converges to a higher value (3.93). A high temperature makes the gate highly permissive, suggesting that in early stages where features are low-level (e.g., edges, colors), the model learns to aggregate a wide range of information without aggressive pruning.

In \textbf{Stage 2}, as features become more semantically meaningful, the temperature drops significantly to 0.68. This makes the gate more selective, allowing the model to form sparser, more content-aware connections. Finally, in \textbf{Stage 3}, the temperature stabilizes at 1.11, very near its initialization value of 1.0 detailed in Section \ref{app:hyperparams}. This suggests that a temperature around 1.0 represents a stable, optimal trade-off for gating the highly abstract features in the deepest layers. This trend demonstrates that our gating mechanism adaptively learns the appropriate level of sparsity for features at different semantic levels, validating the robustness of our approach.

\subsection{Ablation on Gating Function Design}
\label{app:ablation_gating_func}

To justify our choice of using the L1 distance and an exponential decay function for our gate, we compare it against other plausible designs in Table~\ref{tab:ablation_gating_func}. Using the L2 distance, which is more sensitive to outliers, results in a 0.1\% accuracy drop. Replacing the exponential decay with a standard sigmoid gate leads to a more significant 0.3\% drop. These results confirm that our proposed combination of L1 distance for robust feature comparison and exponential decay for stable gating is the most effective design.

\begin{table}[ht]
\def\arraystretch{1.2}
\small
\caption{{\textbf{Ablation on the gating function design} in AdaptViG-B. Our L1 + Exponential Decay design is superior.}}
\centering
\begin{tabular}[t]{|l|l|c|}
\hline
{\textbf{Distance Metric}} & {\textbf{Gating Function}} & {\textbf{Top-1 (\%)}} \\ \hline
{L2 Norm} & {Exponential Decay} & {83.2} \\
{L1 Norm} & {Sigmoid} & {83.0} \\
\rowcolor{Gray}
{\textbf{L1 Norm (Ours)}} & {\textbf{Exponential Decay (Ours)}} & {\textbf{83.3}} \\ \hline
\end{tabular}
\label{tab:ablation_gating_func}
\end{table}

\subsection{Ablation on Feature Fusion Strategy}
\label{app:ablation_fusion}

Our AGC module fuses information from neighbors using a max-relative difference operation following prior work \cite{MobileViG, MobileViGv2}. In Table~\ref{tab:ablation_fusion}, we compare this against several common alternatives: simple addition of neighbor features, channel-wise concatenation, and sum aggregation used in Graph Isomorphism Networks (GIN)~\cite{xu2019how}. Relying on simple addition leads to a 0.5\% drop in accuracy, likely due to feature saturation. Using only concatenation performs better but is still 0.4\% worse than our method. Comparing to sum aggregation from GINs \cite{xu2019how} our approach yields a +0.2\% improvement. This validates that the max-relative difference operation is the most effective fusion strategy for our architecture.

\begin{table}[ht]
\def\arraystretch{1.2}
\caption{\textbf{Ablation on the feature fusion strategy} in AdaptViG-B, following the graph convolution step.}
\centering
\begin{tabular}[t]{|l|c|}
\hline
{\textbf{Fusion Strategy}} & {\textbf{Top-1 (\%)}} \\ \hline
{Simple Addition} & {82.8} \\
{Concatenation} & {82.9} \\
{GIN (Sum Aggregation)~\cite{xu2019how}} & {83.1} \\
\rowcolor{Gray}
{\textbf{Max-Relative Difference (Ours)}} & {\textbf{83.3}} \\ \hline
\end{tabular}
\label{tab:ablation_fusion}
\end{table}

\newpage
\clearpage

\section{Implementation Details}
\label{Sec:Suppl_configuration}

\subsection{Network Configuration}

The detailed network architectures for AdaptViG-S, M, and B are provided in Table~\ref{table:adaptvig_arch}. We report the configuration of the stem, the four stages, and the final classification head. In each stage, we list the number of local Inverted Residual Blocks (IRBs) and global processing blocks (either AGC or Attention), as well as the channel dimensions ($C$).

\begin{table}[H]
\scriptsize
\caption{\textbf{Architecture details of AdaptViG.} showing configuration of the stem, stages, and classification head. $C$ represents the channel dimensions.}
\centering
\setlength{\tabcolsep}{1pt}
\begin{tabular}{|c|c|c|c|}
\hline
\textbf{Stage} & \textbf{AdaptViG-S} & \textbf{AdaptViG-M} & \textbf{AdaptViG-B} \\ \hline \rule{0pt}{4ex}
Stem & Conv $\times$2, & Conv $\times$2, & Conv $\times$2, \\[8pt] \hline \rule{0pt}{4ex}
Stage 1 & $ \begin{array}{c} \text{IRB Block} \times 3 \\ \text{AGC Block} \times 3 \\ C = 32 \end{array} $ & $ \begin{array}{c} \text{IRB Block} \times 4 \\ \text{AGC Block} \times 4 \\ C = 48 \end{array} $ 
& $ \begin{array}{c} \text{IRB Block} \times 5 \\ \text{AGC Block} \times 5 \\ C = 48 \end{array} $ 
\\[8pt] \hline \rule{0pt}{4ex}
Stage 2 & $ \begin{array}{c} \text{IRB Block} \times 3 \\ \text{AGC Block} \times 3 \\ C = 64 \end{array} $ & $ \begin{array}{c} \text{IRB Block} \times 4 \\ \text{AGC Block} \times 4 \\ C = 96 \end{array} $ 
& $ \begin{array}{c} \text{IRB Block} \times 5 \\ \text{AGC Block} \times 5 \\ C = 96 \end{array} $
\\[8pt] \hline \rule{0pt}{4ex}
Stage 3 & $ \begin{array}{c} \text{IRB Block} \times 9 \\ \text{AGC Block} \times 3 \\ C = 128 \end{array} $ & $ \begin{array}{c} \text{IRB Block} \times 12 \\ \text{AGC Block} \times 4 \\ C = 192 \end{array} $ 
& $ \begin{array}{c} \text{IRB Block} \times 15 \\ \text{AGC Block} \times 5 \\ C = 192 \end{array} $
\\[8pt] \hline \rule{0pt}{4ex}
Stage 4 & $ \begin{array}{c} \text{IRB Block} \times 3 \\ \text{Attention Block} \times 3 \\ C = 256 \end{array} $ & $ \begin{array}{c} \text{IRB Block} \times 4 \\ \text{Attention Block} \times 4 \\ C = 320 \end{array} $ 
& $ \begin{array}{c} \text{IRB Block} \times 5 \\ \text{Attention Block} \times 5 \\ C = 384 \end{array} $ \\[8pt] \hline 
Head & \multicolumn{3}{c|}{Pooling \& MLP} \\ \hline
\end{tabular}
\label{table:adaptvig_arch}
\end{table}

\subsection{Hyperparameter Settings}
\label{app:hyperparams}

The detailed hyperparameter settings used for our ImageNet-1K training are provided in Table~\ref{tab:supp_hyperparams}. The hyperparameter settings match those of Vision GNN \cite{Vision_GNN}, MobileViG \cite{MobileViG}, and EfficientFormer \cite{EfficientFormer} for fair comparison.

\begin{table}[ht]
\centering
\caption{\textbf{Training hyperparameters for ImageNet-1K.}}
\def\arraystretch{1.1}
\resizebox{0.825\columnwidth}{!}{%
\begin{tabular}{lc}
\toprule
\textbf{Hyperparameter} & \textbf{Value} \\
\midrule
{Epochs} & 300 \\
{Optimizer} & AdamW \cite{AdamW} \\
{Batch Size} & 1024 \\
{Start Learning Rate (LR)} & 2e$^{-3}$ \\
{LR Schedule} & Cosine \\
{Warmup Epochs} & 20 \\
{Weight Decay} & 0.05 \\
{Repeated Augment \cite{RepeatedAugment}} & {\checkmark} \\
{RandAugment \cite{RandAugment}} & {\checkmark} \\
{Mixup Prob. \cite{Mixup}} & {0.8} \\
{Cutmix Prob. \cite{CutMix}} & {1.0} \\
{Random Erasing Prob. \cite{RandomErase}} & {0.25} \\
{Exponential Moving Average} & {0.99996} \\
{Temperature (T) Initialization} & {1.0} \\
\bottomrule
\end{tabular}
}
\label{tab:supp_hyperparams}
\end{table}

\newpage

\section{Additional Experimental Results}
\label{Sec:Suppl_Experimental}

\subsection{CIFAR-100 Image Classification Results}
\label{supp:Cifar100}

We conduct image classification experiments on the CIFAR-100~\cite{Cifar} dataset, training from scratch for 200 epochs. We report the top-1 accuracy on the test set and implement all models using PyTorch~\cite{paszke2019pytorch} and the Timm library~\cite{timm} with the AdamW~\cite{AdamW} optimizer and a cosine annealing schedule.

\begin{table}[!htb]
\def\arraystretch{1.2}
\caption{\textbf{Results of our AdaptViG-S and competing methods on the CIFAR-100} image classification task.}
\footnotesize
\centering
\begin{tabular}[t]{l|c|c|c}
\hline
\textbf{Model} & \textbf{Type} & \textbf{Params (M)} & \textbf{Top-1 (\%)} \\ \hline
ResNet-50~\cite{Resnet} & CNN & 23.7 & 80.9 \\
ConvNeXt-T~\cite{liu2022convnet} & CNN & 28.0 & 82.5 \\
MobileViG-Ti~\cite{MobileViG} & CNN-GNN & 4.3 & 80.2 \\
MobileViG-B~\cite{MobileViG} & CNN-GNN & 25.4 & 83.8 \\
Swin-T~\cite{liu2021swin} & ViT & 28.0 & 74.9 \\
\rowcolor{Gray}
\textbf{AdaptViG-S (Ours)} & \textbf{CNN-GNN} & \textbf{7.5} & \textbf{84.0} \\ \hline
\end{tabular}
\label{tab:class_Cifar}
\end{table}

As shown in Table~\ref{tab:class_Cifar}, our AdaptViG-S model achieves state-of-the-art performance among efficient models on CIFAR-100. With only 7.5M parameters, it obtains 84.0\% Top-1 accuracy, outperforming MobileViG-B by 0.2\% while using 71\% fewer parameters, demonstrating its superior efficiency.

\subsection{CIFAR-10 Image Classification Results}
\label{supp:Cifar10}

We conduct further image classification experiments on the CIFAR-10~\cite{Cifar} dataset, which consists of 10 object classes, training from scratch for 200 epochs.

\begin{table}[h]
\def\arraystretch{1.2}
\caption{\textbf{Results of our AdaptViG-S and competing methods on the on the CIFAR-10} image classification task.}
\footnotesize
\centering
\begin{tabular}[t]{l|c|c|c}
\hline
\textbf{Model} & \textbf{Type} & \textbf{Params (M)} & \textbf{Top-1 Acc (\%)} \\ \hline
ConvNeXt-T~\cite{liu2022convnet} & CNN & 28.0 & 97.1 \\
MobileViG-Ti~\cite{MobileViG} & CNN-GNN & 4.3 & 95.6 \\
MobileViG-B \cite{MobileViG}   & CNN-GNN & 25.3        & 96.7 \\
Swin-T~\cite{liu2021swin} & ViT & 28.0 & 91.1 \\
\rowcolor{Gray}
\textbf{AdaptViG-S (Ours)} & \textbf{CNN-GNN} & \textbf{7.5} & \textbf{97.0} \\ \hline
\end{tabular}
\label{tab:class_Cifar10}
\end{table}

When tested on CIFAR-10, as shown in Table~\ref{tab:class_Cifar10}, AdaptViG-S continues to show superior performance. It achieves 97.0\% Top-1 accuracy, which is 5.9\% higher than Swin-T with 73\% fewer parameters. Our model also achieves virtually the same state-of-the-art performance as ConvNeXt-T (97.0\% vs. 97.1\%) while using only a fraction of the parameters (7.5M vs. 28.0M), a 73\% reduction.

\subsection{Medical Image Classification Results}
\label{supp:medical_results}
We evaluate AdaptViG on two distinct medical image classification tasks from the MedMNISTv2 benchmark~\cite{yang2023medmnist}. The first, OrganSMNIST, consists of 11 organ classes from 13,932 training and 2,452 validation abdominal CT images. The second, DermaMNIST, consists of 7 skin lesion classes from 7,007 training and 1,003 validation Dermatoscope images. For both datasets, we train models from scratch for 200 epochs using the AdamW optimizer~\cite{AdamW} and a cosine annealing schedule. All implementations use PyTorch~\cite{paszke2019pytorch} and the Timm library~\cite{timm}.

\begin{table}[h]
\def\arraystretch{1.2}
\caption{\textbf{Results on the OrganSMNIST} medical image classification task.}
\footnotesize
\centering
\begin{tabular}[t]{l|c|c|c}
\hline
\textbf{Model} & \textbf{Type} & \textbf{Params (M)} & \textbf{Top-1 Acc (\%)} \\
\hline
ResNet-50~\cite{Resnet} & CNN & 23.7 & 92.6 \\
ConvNeXt-T~\cite{liu2022convnet} & CNN & 28.0 & 91.6 \\
MobileViG-Ti~\cite{MobileViG} & CNN-GNN & 4.3 & 90.7 \\
Swin-T~\cite{liu2021swin} & ViT & 28.0 & 91.7 \\
\rowcolor{Gray}
\textbf{AdaptViG-S (Ours)} & \textbf{CNN-GNN} & \textbf{7.5} & \textbf{92.0} \\
\hline
\end{tabular}
\label{tab:med_cls_organs_mnist}
\end{table}

\vspace{3mm}

\paragraph{OrganSMNIST.}
On the OrganSMNIST dataset, as shown in Table~\ref{tab:med_cls_organs_mnist}, our AdaptViG-S model demonstrates superior performance and efficiency. It achieves a Top-1 accuracy of \textbf{92.0\%}, which is higher than both Swin-T~\cite{liu2021swin} (+0.3\%) and ConvNeXt-T~\cite{liu2022convnet} (+0.4\%) while using approximately 73\% fewer parameters than both. This result highlights our model's ability to learn effective representations for volumetric medical data.

\begin{table}[h]
\def\arraystretch{1.2}
\caption{\textbf{Results on the DermaMNIST} medical image classification task.}
\footnotesize
\centering
\begin{tabular}[t]{l|c|c|c}
\hline
\textbf{Model} & \textbf{Type} & \textbf{Params (M)} & \textbf{Top-1 Acc (\%)} \\
\hline
ResNet-50~\cite{Resnet} & CNN & 23.7 & 76.1 \\
ConvNeXt-T~\cite{liu2022convnet} & CNN & 28.0 & 78.1 \\
MobileViG-Ti~\cite{MobileViG} & CNN-GNN & 4.3 & 75.4 \\
Swin-T~\cite{liu2021swin} & ViT & 28.0 & 80.0 \\
\rowcolor{Gray}
\textbf{AdaptViG-S (Ours)} & \textbf{CNN-GNN} & \textbf{7.5} & \textbf{76.1} \\
\hline
\end{tabular}
\label{tab:med_cls_derma_mnist}
\end{table}

\vspace{3mm}

\paragraph{DermaMNIST.}
When tested on the DermaMNIST dataset, our AdaptViG-S continues to show a strong efficiency profile. As shown in Table~\ref{tab:med_cls_derma_mnist}, it achieves 76.0\% accuracy, outperforming the lighter MobileViG-Ti \cite{MobileViG} by 0.7\%. While Swin-T \cite{liu2021swin} and ConvNeXt-T \cite{liu2022convnet} achieve higher absolute accuracy, our model provides competitive performance with significantly fewer parameters, underscoring its utility in resource-constrained medical imaging applications.

\clearpage

\section{Qualitative Visualizations}

\begin{figure}[htbp]
    \centering
    \begin{subfigure}[b]{0.48\textwidth}
        \centering
        \includegraphics[width=\textwidth]{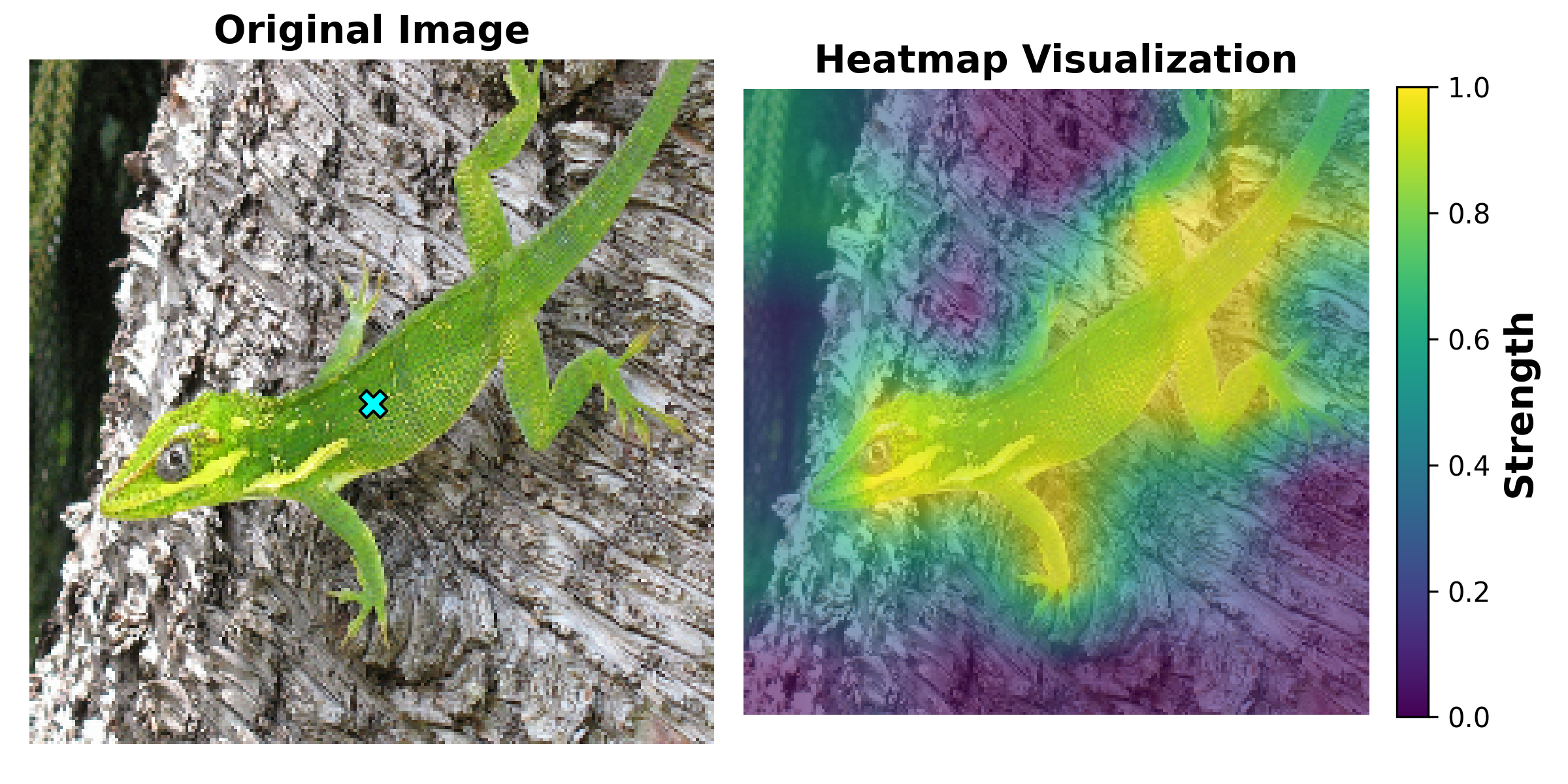}
        \caption{Standard case with a clear object and background.}
    \end{subfigure}
    
    \vspace{5mm}
    
    \begin{subfigure}[b]{0.48\textwidth}
        \centering
        \includegraphics[width=\textwidth]{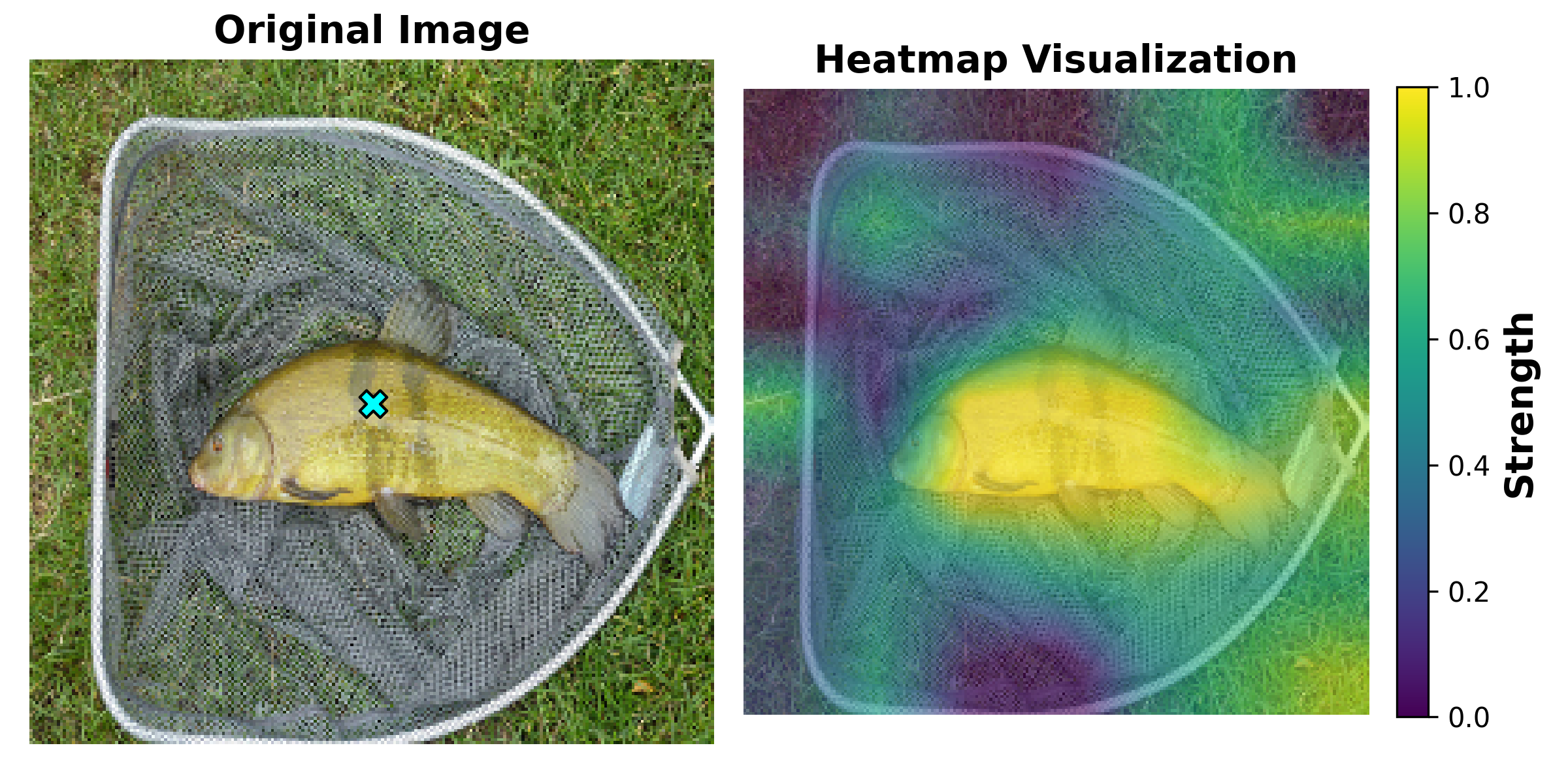}
        \caption{Challenging case with similar color object and background.}
    \end{subfigure}
    \caption{{\textbf{Visualization of learned spatial interaction strength.} From a reference point on the primary object (cyan 'X'), we visualize the learned weights the model assigns across the image. Warmer colors (yellow) indicate higher interaction strength. The model correctly focuses on the object in both a standard case (lizard) and a more challenging one with similar object and background colors (fish), demonstrating its ability to learn robust, object-aware features.}}
    \label{fig:qual_viz}
\end{figure}

To provide insight into our model's learned behavior, we perform a qualitative analysis by visualizing the learned interaction strength within one of the model's global blocks. In Figure~\ref{fig:qual_viz}, for a selected reference point on the primary object (marked with a cyan 'X'), we compute and overlay a heatmap of the learned spatial weights across the image.

The visualizations demonstrate that our model learns to focus on semantically coherent regions, validating our design. In the standard case (a), the reference point on the lizard's body yields high interaction scores across the lizard's entire body, including its head and tail, while correctly suppressing the distinct tree background. In the more challenging case (b), where the fish's color is similar to the surrounding grass, the model still demonstrates a strong ability to focus on the coherent shape of the fish. This shows that our architecture learns robust feature representations that can identify objects from their backgrounds.

\end{document}


%% file: main.bbl
\begin{thebibliography}{68}
\providecommand{\natexlab}[1]{#1}
\providecommand{\url}[1]{\texttt{#1}}
\expandafter\ifx\csname urlstyle\endcsname\relax
  \providecommand{\doi}[1]{doi: #1}\else
  \providecommand{\doi}{doi: \begingroup \urlstyle{rm}\Url}\fi

\bibitem[Alon(1986)]{spectral_gap_eigenvalues}
Noga Alon.
\newblock Eigenvalues and expanders.
\newblock \emph{Combinatorica}, 6\penalty0 (2):\penalty0 83--96, 1986.

\bibitem[Alon and Yahav(2020)]{alon2020bottleneck}
Uri Alon and Eran Yahav.
\newblock On the bottleneck of graph neural networks and its practical implications.
\newblock \emph{arXiv preprint arXiv:2006.05205}, 2020.

\bibitem[Avery et~al.(2024)Avery, Munir, and Marculescu]{MobileViGv2}
William Avery, Mustafa Munir, and Radu Marculescu.
\newblock Scaling graph convolutions for mobile vision.
\newblock In \emph{Proceedings of the IEEE/CVF Conference on Computer Vision and Pattern Recognition (CVPR) Workshops}, pages 5857--5865, 2024.

\bibitem[Ba et~al.(2016)Ba, Kiros, and Hinton]{LayerNorm}
Jimmy~Lei Ba, Jamie~Ryan Kiros, and Geoffrey~E Hinton.
\newblock Layer normalization.
\newblock \emph{arXiv preprint arXiv:1607.06450}, 2016.

\bibitem[Chen et~al.(2024)Chen, Wu, Dai, Zhou, Xu, Yang, Han, and Yu]{chen2024survey}
Chaoqi Chen, Yushuang Wu, Qiyuan Dai, Hong-Yu Zhou, Mutian Xu, Sibei Yang, Xiaoguang Han, and Yizhou Yu.
\newblock A survey on graph neural networks and graph transformers in computer vision: A task-oriented perspective.
\newblock \emph{IEEE Transactions on Pattern Analysis and Machine Intelligence}, 2024.

\bibitem[Chen et~al.(2021)Chen, Fan, and Panda]{CrossViT}
Chun-Fu~Richard Chen, Quanfu Fan, and Rameswar Panda.
\newblock Crossvit: Cross-attention multi-scale vision transformer for image classification.
\newblock In \emph{Proceedings of the IEEE/CVF international conference on computer vision}, pages 357--366, 2021.

\bibitem[Chen et~al.(2022)Chen, Dai, Chen, Liu, Dong, Yuan, and Liu]{MobileFormer}
Yinpeng Chen, Xiyang Dai, Dongdong Chen, Mengchen Liu, Xiaoyi Dong, Lu Yuan, and Zicheng Liu.
\newblock Mobile-former: Bridging mobilenet and transformer.
\newblock In \emph{Proceedings of the IEEE/CVF Conference on Computer Vision and Pattern Recognition}, pages 5270--5279, 2022.

\bibitem[Chu et~al.(2021)Chu, Tian, Zhang, Wang, Wei, Xia, and Shen]{CPE}
Xiangxiang Chu, Zhi Tian, Bo Zhang, Xinlong Wang, Xiaolin Wei, Huaxia Xia, and Chunhua Shen.
\newblock Conditional positional encodings for vision transformers.
\newblock \emph{arXiv preprint arXiv:2102.10882}, 2021.

\bibitem[Cover and Hart(1967)]{KNN}
Thomas Cover and Peter Hart.
\newblock Nearest neighbor pattern classification.
\newblock \emph{IEEE transactions on information theory}, 13\penalty0 (1):\penalty0 21--27, 1967.

\bibitem[Cubuk et~al.(2020)Cubuk, Zoph, Shlens, and Le]{RandAugment}
Ekin~D Cubuk, Barret Zoph, Jonathon Shlens, and Quoc~V Le.
\newblock Randaugment: Practical automated data augmentation with a reduced search space.
\newblock In \emph{Proceedings of the IEEE/CVF Conference on Computer Vision and Pattern Recognition Workshops}, pages 702--703, 2020.

\bibitem[Deng et~al.(2009)Deng, Dong, Socher, Li, Li, and Fei-Fei]{imagenet1k}
Jia Deng, Wei Dong, Richard Socher, Li-Jia Li, Kai Li, and Li Fei-Fei.
\newblock Imagenet: A large-scale hierarchical image database.
\newblock In \emph{2009 IEEE Conference on Computer Vision and Pattern Recognition}, pages 248--255, 2009.

\bibitem[Dosovitskiy et~al.(2020)]{ViT}
Alexey Dosovitskiy et~al.
\newblock An image is worth 16x16 words: Transformers for image recognition at scale.
\newblock \emph{arXiv preprint arXiv:2010.11929}, 2020.

\bibitem[Graham et~al.(2021)Graham, El-Nouby, Touvron, Stock, Joulin, J{\'e}gou, and Douze]{graham2021levit}
Benjamin Graham, Alaaeldin El-Nouby, Hugo Touvron, Pierre Stock, Armand Joulin, Herv{\'e} J{\'e}gou, and Matthijs Douze.
\newblock Levit: a vision transformer in convnet's clothing for faster inference.
\newblock In \emph{Proceedings of the IEEE/CVF international Conference on Computer Vision}, pages 12259--12269, 2021.

\bibitem[Gu and Dao(2023)]{Mamba}
Albert Gu and Tri Dao.
\newblock Mamba: Linear-time sequence modeling with selective state spaces.
\newblock \emph{arXiv preprint arXiv:2312.00752}, 2023.

\bibitem[Han et~al.(2022)Han, Wang, Guo, Tang, and Wu]{Vision_GNN}
Kai Han, Yunhe Wang, Jianyuan Guo, Yehui Tang, and Enhua Wu.
\newblock Vision gnn: An image is worth graph of nodes.
\newblock \emph{arXiv preprint arXiv:2206.00272}, 2022.

\bibitem[Han et~al.(2023)Han, Wang, Kundu, Ding, and Wang]{ViHGNN}
Yan Han, Peihao Wang, Souvik Kundu, Ying Ding, and Zhangyang Wang.
\newblock Vision hgnn: An image is more than a graph of nodes.
\newblock In \emph{Proceedings of the IEEE/CVF International Conference on Computer Vision}, pages 19878--19888, 2023.

\bibitem[He et~al.(2016)He, Zhang, Ren, and Sun]{Resnet}
Kaiming He, Xiangyu Zhang, Shaoqing Ren, and Jian Sun.
\newblock Deep residual learning for image recognition.
\newblock In \emph{Proceedings of the IEEE/CVF Conference on Computer Vision and Pattern Recognition}, pages 770--778, 2016.

\bibitem[He et~al.(2017)He, Gkioxari, Doll{\'a}r, and Girshick]{mask_r_cnn}
Kaiming He, Georgia Gkioxari, Piotr Doll{\'a}r, and Ross Girshick.
\newblock Mask r-cnn.
\newblock In \emph{Proceedings of the IEEE International Conference on Computer Vision}, pages 2961--2969, 2017.

\bibitem[Hoffer et~al.(2020)Hoffer, Ben-Nun, Hubara, Giladi, Hoefler, and Soudry]{RepeatedAugment}
Elad Hoffer, Tal Ben-Nun, Itay Hubara, Niv Giladi, Torsten Hoefler, and Daniel Soudry.
\newblock Augment your batch: Improving generalization through instance repetition.
\newblock In \emph{Proceedings of the IEEE/CVF Conference on Computer Vision and Pattern Recognition}, pages 8129--8138, 2020.

\bibitem[Howard et~al.(2017)Howard, Zhu, Chen, Kalenichenko, Wang, Weyand, Andreetto, and Adam]{MobileNet}
Andrew~G Howard, Menglong Zhu, Bo Chen, Dmitry Kalenichenko, Weijun Wang, Tobias Weyand, Marco Andreetto, and Hartwig Adam.
\newblock Mobilenets: Efficient convolutional neural networks for mobile vision applications.
\newblock \emph{arXiv preprint arXiv:1704.04861}, 2017.

\bibitem[Huang et~al.(2023)Huang, Aviles-Rivero, Sch{\"o}nlieb, and Yang]{huang2023vigu}
Jiahao Huang, Angelica~I Aviles-Rivero, Carola-Bibiane Sch{\"o}nlieb, and Guang Yang.
\newblock Vigu: Vision gnn u-net for fast mri.
\newblock In \emph{2023 IEEE 20th International Symposium on Biomedical Imaging (ISBI)}, pages 1--5. IEEE, 2023.

\bibitem[Jiao et~al.(2022)Jiao, Chen, Liu, Yang, You, Liu, Li, and Hou]{jiao2022graph}
Licheng Jiao, Jie Chen, Fang Liu, Shuyuan Yang, Chao You, Xu Liu, Lingling Li, and Biao Hou.
\newblock Graph representation learning meets computer vision: A survey.
\newblock \emph{IEEE Transactions on Artificial Intelligence}, 4\penalty0 (1):\penalty0 2--22, 2022.

\bibitem[Kingma and Ba(2014)]{Adam}
Diederik~P Kingma and Jimmy Ba.
\newblock Adam: A method for stochastic optimization.
\newblock \emph{arXiv preprint arXiv:1412.6980}, 2014.

\bibitem[Kirillov et~al.(2019)Kirillov, Girshick, He, and Doll{\'a}r]{kirillov2019panoptic}
Alexander Kirillov, Ross Girshick, Kaiming He, and Piotr Doll{\'a}r.
\newblock Panoptic feature pyramid networks.
\newblock In \emph{Proceedings of the IEEE/CVF Conference on Computer Vision and Pattern Recognition}, pages 6399--6408, 2019.

\bibitem[Krizhevsky et~al.(2009)Krizhevsky, Hinton, et~al.]{Cifar}
Alex Krizhevsky, Geoffrey Hinton, et~al.
\newblock Learning multiple layers of features from tiny images.
\newblock 2009.

\bibitem[Krizhevsky et~al.(2012)Krizhevsky, Sutskever, and Hinton]{Alexnet2012}
Alex Krizhevsky, Ilya Sutskever, and Geoffrey~E Hinton.
\newblock Imagenet classification with deep convolutional neural networks.
\newblock \emph{Advances in Neural Information Processing Systems}, 2012.

\bibitem[Li et~al.(2019)Li, Muller, Thabet, and Ghanem]{maxrel}
Guohao Li, Matthias Muller, Ali Thabet, and Bernard Ghanem.
\newblock Deepgcns: Can gcns go as deep as cnns?
\newblock In \emph{Proceedings of the IEEE/CVF International Conference on Computer Vision}, pages 9267--9276, 2019.

\bibitem[Li et~al.(2022{\natexlab{a}})Li, Xia, Li, Li, Wang, Xiao, Wang, Zheng, and Pan]{li2022next}
Jiashi Li, Xin Xia, Wei Li, Huixia Li, Xing Wang, Xuefeng Xiao, Rui Wang, Min Zheng, and Xin Pan.
\newblock Next-vit: Next generation vision transformer for efficient deployment in realistic industrial scenarios.
\newblock \emph{arXiv preprint arXiv:2207.05501}, 2022{\natexlab{a}}.

\bibitem[Li et~al.(2022{\natexlab{b}})Li, Wang, Liu, Tan, Lin, Wu, Chen, Zheng, and Li]{MogaNet}
Siyuan Li, Zedong Wang, Zicheng Liu, Cheng Tan, Haitao Lin, Di Wu, Zhiyuan Chen, Jiangbin Zheng, and Stan~Z Li.
\newblock Moganet: Multi-order gated aggregation network.
\newblock \emph{arXiv preprint arXiv:2211.03295}, 2022{\natexlab{b}}.

\bibitem[Li et~al.(2022{\natexlab{c}})Li, Hu, Wen, Evangelidis, Salahi, Wang, Tulyakov, and Ren]{EfficientFormerv2}
Yanyu Li, Ju Hu, Yang Wen, Georgios Evangelidis, Kamyar Salahi, Yanzhi Wang, Sergey Tulyakov, and Jian Ren.
\newblock Rethinking vision transformers for mobilenet size and speed.
\newblock \emph{arXiv preprint arXiv:2212.08059}, 2022{\natexlab{c}}.

\bibitem[Li et~al.(2022{\natexlab{d}})Li, Yuan, Wen, Hu, Evangelidis, Tulyakov, Wang, and Ren]{EfficientFormer}
Yanyu Li, Geng Yuan, Yang Wen, Ju Hu, Georgios Evangelidis, Sergey Tulyakov, Yanzhi Wang, and Jian Ren.
\newblock Efficientformer: Vision transformers at mobilenet speed.
\newblock \emph{Advances in Neural Information Processing Systems}, 35:\penalty0 12934--12949, 2022{\natexlab{d}}.

\bibitem[Lin et~al.(2014)Lin, Maire, Belongie, Hays, Perona, Ramanan, Doll{\'a}r, and Zitnick]{coco}
Tsung-Yi Lin, Michael Maire, Serge Belongie, James Hays, Pietro Perona, Deva Ramanan, Piotr Doll{\'a}r, and C~Lawrence Zitnick.
\newblock Microsoft coco: Common objects in context.
\newblock In \emph{Proceedings of the European Conference on Computer Vision}, pages 740--755. Springer, 2014.

\bibitem[Liu et~al.(2024)Liu, Tian, Zhao, Yu, Xie, Wang, Ye, Jiao, and Liu]{VMamba}
Yue Liu, Yunjie Tian, Yuzhong Zhao, Hongtian Yu, Lingxi Xie, Yaowei Wang, Qixiang Ye, Jianbin Jiao, and Yunfan Liu.
\newblock Vmamba: Visual state space model.
\newblock \emph{Advances in neural information processing systems}, 37:\penalty0 103031--103063, 2024.

\bibitem[Liu et~al.(2021)Liu, Lin, Cao, Hu, Wei, Zhang, Lin, and Guo]{liu2021swin}
Ze Liu, Yutong Lin, Yue Cao, Han Hu, Yixuan Wei, Zheng Zhang, Stephen Lin, and Baining Guo.
\newblock Swin transformer: Hierarchical vision transformer using shifted windows.
\newblock In \emph{Proceedings of the IEEE/CVF International Conference on Computer Vision}, pages 10012--10022, 2021.

\bibitem[Liu et~al.(2022)Liu, Mao, Wu, Feichtenhofer, Darrell, and Xie]{liu2022convnet}
Zhuang Liu, Hanzi Mao, Chao-Yuan Wu, Christoph Feichtenhofer, Trevor Darrell, and Saining Xie.
\newblock A convnet for the 2020s.
\newblock In \emph{Proceedings of the IEEE/CVF Conference on Computer Vision and Pattern Recognition}, pages 11976--11986, 2022.

\bibitem[Loshchilov and Hutter(2017)]{AdamW}
Ilya Loshchilov and Frank Hutter.
\newblock Decoupled weight decay regularization.
\newblock \emph{arXiv preprint arXiv:1711.05101}, 2017.

\bibitem[Lubotzky et~al.(1988)Lubotzky, Phillips, and Sarnak]{lubotzky1988ramanujan}
Alexander Lubotzky, Ralph Phillips, and Peter Sarnak.
\newblock Ramanujan graphs.
\newblock \emph{Combinatorica}, 8\penalty0 (3):\penalty0 261--277, 1988.

\bibitem[Mehta and Rastegari(2022{\natexlab{a}})]{MobileViT}
Sachin Mehta and Mohammad Rastegari.
\newblock Mobilevit: light-weight, general-purpose, and mobile-friendly vision transformer.
\newblock In \emph{International Conference on Learning Representations}, 2022{\natexlab{a}}.

\bibitem[Mehta and Rastegari(2022{\natexlab{b}})]{MobileViTv2}
Sachin Mehta and Mohammad Rastegari.
\newblock Separable self-attention for mobile vision transformers.
\newblock \emph{arXiv preprint arXiv:2206.02680}, 2022{\natexlab{b}}.

\bibitem[Mehta et~al.(2018)Mehta, Rastegari, Caspi, Shapiro, and Hajishirzi]{mehta2018espnet}
Sachin Mehta, Mohammad Rastegari, Anat Caspi, Linda Shapiro, and Hannaneh Hajishirzi.
\newblock Espnet: Efficient spatial pyramid of dilated convolutions for semantic segmentation.
\newblock In \emph{Proceedings of the European Conference on Computer Vision}, pages 552--568, 2018.

\bibitem[Munir et~al.(2023)Munir, Avery, and Marculescu]{MobileViG}
Mustafa Munir, William Avery, and Radu Marculescu.
\newblock Mobilevig: Graph-based sparse attention for mobile vision applications.
\newblock In \emph{Proceedings of the IEEE/CVF Conference on Computer Vision and Pattern Recognition Workshops}, pages 2211--2219, 2023.

\bibitem[Munir et~al.(2024{\natexlab{a}})Munir, Avery, Rahman, and Marculescu]{GreedyViG}
Mustafa Munir, William Avery, Md~Mostafijur Rahman, and Radu Marculescu.
\newblock Greedyvig: Dynamic axial graph construction for efficient vision gnns.
\newblock In \emph{Proceedings of the IEEE/CVF Conference on Computer Vision and Pattern Recognition (CVPR)}, pages 6118--6127, 2024{\natexlab{a}}.

\bibitem[Munir et~al.(2024{\natexlab{b}})Munir, Zhang, and Marculescu]{LogViG}
Mustafa Munir, Alex Zhang, and Radu Marculescu.
\newblock Multi-scale high-resolution logarithmic grapher module for efficient vision gnns.
\newblock In \emph{The Third Learning on Graphs Conference}, 2024{\natexlab{b}}.

\bibitem[Parikh et~al.(2025)Parikh, Fein-Ashley, Ye, Kannan, and Prasanna]{parikh2025clustervig}
Dhruv Parikh, Jacob Fein-Ashley, Tian Ye, Rajgopal Kannan, and Viktor Prasanna.
\newblock Clustervig: Efficient globally aware vision gnns via image partitioning.
\newblock \emph{arXiv preprint arXiv:2501.10640}, 2025.

\bibitem[Paszke et~al.(2019)]{paszke2019pytorch}
Adam Paszke et~al.
\newblock Pytorch: An imperative style, high-performance deep learning library.
\newblock \emph{Advances in Neural Information Processing Systems}, 32, 2019.

\bibitem[Rao et~al.(2022)Rao, Zhao, Tang, Zhou, Lim, and Lu]{HorNet}
Yongming Rao, Wenliang Zhao, Yansong Tang, Jie Zhou, Ser~Nam Lim, and Jiwen Lu.
\newblock Hornet: Efficient high-order spatial interactions with recursive gated convolutions.
\newblock \emph{Advances in Neural Information Processing Systems}, 35:\penalty0 10353--10366, 2022.

\bibitem[Sandler et~al.(2018)Sandler, Howard, Zhu, Zhmoginov, and Chen]{MobileNetv2}
Mark Sandler, Andrew Howard, Menglong Zhu, Andrey Zhmoginov, and Liang-Chieh Chen.
\newblock Mobilenetv2: Inverted residuals and linear bottlenecks.
\newblock In \emph{Proceedings of the IEEE conference on computer vision and pattern recognition}, pages 4510--4520, 2018.

\bibitem[Simonyan and Zisserman(2014)]{VGGNet}
Karen Simonyan and Andrew Zisserman.
\newblock Very deep convolutional networks for large-scale image recognition.
\newblock \emph{arXiv preprint arXiv:1409.1556}, 2014.

\bibitem[Spadaro et~al.(2025)Spadaro, Grangetto, Fiandrotti, Tartaglione, and Giraldo]{spadaro2025wignet}
Gabriele Spadaro, Marco Grangetto, Attilio Fiandrotti, Enzo Tartaglione, and Jhony~H Giraldo.
\newblock Wignet: Windowed vision graph neural network.
\newblock In \emph{2025 IEEE/CVF Winter Conference on Applications of Computer Vision (WACV)}, pages 859--868. IEEE, 2025.

\bibitem[Tan and Le(2019)]{tan2019efficientnet}
Mingxing Tan and Quoc Le.
\newblock Efficientnet: Rethinking model scaling for convolutional neural networks.
\newblock In \emph{International Conference on Machine Learning}, pages 6105--6114. PMLR, 2019.

\bibitem[Tan and Le(2021)]{tan2021efficientnetv2}
Mingxing Tan and Quoc Le.
\newblock Efficientnetv2: Smaller models and faster training.
\newblock In \emph{International Conference on Machine Learning}, pages 10096--10106. PMLR, 2021.

\bibitem[Tu et~al.(2022)Tu, Talebi, Zhang, Yang, Milanfar, Bovik, and Li]{MaxViT}
Zhengzhong Tu, Hossein Talebi, Han Zhang, Feng Yang, Peyman Milanfar, Alan Bovik, and Yinxiao Li.
\newblock Maxvit: Multi-axis vision transformer.
\newblock In \emph{European conference on computer vision}, pages 459--479. Springer, 2022.

\bibitem[Vasu et~al.(2022)Vasu, Gabriel, Zhu, Tuzel, and Ranjan]{mobileone2022}
Pavan Kumar~Anasosalu Vasu, James Gabriel, Jeff Zhu, Oncel Tuzel, and Anurag Ranjan.
\newblock An improved one millisecond mobile backbone.
\newblock \emph{arXiv preprint arXiv:2206.04040}, 2022.

\bibitem[Vasu et~al.(2023)Vasu, Gabriel, Zhu, Tuzel, and Ranjan]{FastViT}
Pavan Kumar~Anasosalu Vasu, James Gabriel, Jeff Zhu, Oncel Tuzel, and Anurag Ranjan.
\newblock Fastvit: A fast hybrid vision transformer using structural reparameterization.
\newblock In \emph{Proceedings of the IEEE/CVF International Conference on Computer Vision}, 2023.

\bibitem[Vaswani et~al.(2017)Vaswani, Shazeer, Parmar, Uszkoreit, Jones, Gomez, Kaiser, and Polosukhin]{vaswani2017attention}
Ashish Vaswani, Noam Shazeer, Niki Parmar, Jakob Uszkoreit, Llion Jones, Aidan~N Gomez, {\L}ukasz Kaiser, and Illia Polosukhin.
\newblock Attention is all you need.
\newblock \emph{Advances in Neural Information Processing Systems}, 30, 2017.

\bibitem[Wang et~al.(2021)Wang, Xie, Li, Fan, Song, Liang, Lu, Luo, and Shao]{wang2021pyramid}
Wenhai Wang, Enze Xie, Xiang Li, Deng-Ping Fan, Kaitao Song, Ding Liang, Tong Lu, Ping Luo, and Ling Shao.
\newblock Pyramid vision transformer: A versatile backbone for dense prediction without convolutions.
\newblock In \emph{Proceedings of the IEEE/CVF International Conference on Computer Vision}, pages 568--578, 2021.

\bibitem[Watts and Strogatz(1998)]{watts1998collective}
Duncan~J Watts and Steven~H Strogatz.
\newblock Collective dynamics of ‘small-world’networks.
\newblock \emph{nature}, 393\penalty0 (6684):\penalty0 440--442, 1998.

\bibitem[Wightman(2019)]{timm}
Ross Wightman.
\newblock {PyTorch Image Models}.
\newblock \url{https://github.com/rwightman/pytorch-image-models}, 2019.

\bibitem[Wu et~al.(2023)Wu, Li, Zhang, Zhang, Chi, Wang, and Wang]{PVG}
JiaFu Wu, Jian Li, Jiangning Zhang, Boshen Zhang, Mingmin Chi, Yabiao Wang, and Chengjie Wang.
\newblock Pvg: Progressive vision graph for vision recognition.
\newblock In \emph{Proceedings of the 31st ACM International Conference on Multimedia}, page 2477–2486, New York, NY, USA, 2023. Association for Computing Machinery.

\bibitem[Xu et~al.(2019)Xu, Hu, Leskovec, and Jegelka]{xu2019how}
Keyulu Xu, Weihua Hu, Jure Leskovec, and Stefanie Jegelka.
\newblock How powerful are graph neural networks?
\newblock In \emph{International Conference on Learning Representations}, 2019.

\bibitem[Yang et~al.(2023)Yang, Shi, Wei, Liu, Zhao, Ke, Pfister, and Ni]{yang2023medmnist}
Jiancheng Yang, Rui Shi, Donglai Wei, Zequan Liu, Lin Zhao, Bilian Ke, Hanspeter Pfister, and Bingbing Ni.
\newblock Medmnist v2-a large-scale lightweight benchmark for 2d and 3d biomedical image classification.
\newblock \emph{Scientific Data}, 10\penalty0 (1):\penalty0 41, 2023.

\bibitem[Yu and Wang(2025)]{MambaOut}
Weihao Yu and Xinchao Wang.
\newblock Mambaout: Do we really need mamba for vision?
\newblock In \emph{Proceedings of the Computer Vision and Pattern Recognition Conference}, pages 4484--4496, 2025.

\bibitem[Yu et~al.(2022)Yu, Luo, Zhou, Si, Zhou, Wang, Feng, and Yan]{MetaFormer}
Weihao Yu, Mi Luo, Pan Zhou, Chenyang Si, Yichen Zhou, Xinchao Wang, Jiashi Feng, and Shuicheng Yan.
\newblock Metaformer is actually what you need for vision.
\newblock In \emph{Proceedings of the IEEE/CVF Conference on Computer Vision and Pattern Recognition}, pages 10819--10829, 2022.

\bibitem[Yun et~al.(2019)Yun, Han, Oh, Chun, Choe, and Yoo]{CutMix}
Sangdoo Yun, Dongyoon Han, Seong~Joon Oh, Sanghyuk Chun, Junsuk Choe, and Youngjoon Yoo.
\newblock Cutmix: Regularization strategy to train strong classifiers with localizable features.
\newblock In \emph{Proceedings of the IEEE/CVF International Conference on Computer Vision}, pages 6023--6032, 2019.

\bibitem[Zhang et~al.(2018)Zhang, Cisse, Dauphin, and Lopez-Paz]{Mixup}
Hongyi Zhang, Moustapha Cisse, Yann~N. Dauphin, and David Lopez-Paz.
\newblock mixup: Beyond empirical risk minimization.
\newblock In \emph{International Conference on Learning Representations}, 2018.

\bibitem[Zhong et~al.(2020)Zhong, Zheng, Kang, Li, and Yang]{RandomErase}
Zhun Zhong, Liang Zheng, Guoliang Kang, Shaozi Li, and Yi Yang.
\newblock Random erasing data augmentation.
\newblock In \emph{Proceedings of the AAAI conference on artificial intelligence}, pages 13001--13008, 2020.

\bibitem[Zhou et~al.(2017)Zhou, Zhao, Puig, Fidler, Barriuso, and Torralba]{ADE20K}
Bolei Zhou, Hang Zhao, Xavier Puig, Sanja Fidler, Adela Barriuso, and Antonio Torralba.
\newblock Scene parsing through ade20k dataset.
\newblock In \emph{Proceedings of the IEEE/CVF Conference on Computer Vision and Pattern Recognition}, pages 633--641, 2017.

\bibitem[Zhu et~al.(2024)Zhu, Liao, Zhang, Wang, Liu, and Wang]{Vim}
Lianghui Zhu, Bencheng Liao, Qian Zhang, Xinlong Wang, Wenyu Liu, and Xinggang Wang.
\newblock Vision mamba: Efficient visual representation learning with bidirectional state space model.
\newblock \emph{arXiv preprint arXiv:2401.09417}, 2024.

\end{thebibliography}
